\documentclass{article}

\usepackage[square,sort,comma,numbers]{natbib}
\usepackage[preprint]{neurips_2021}

\usepackage[utf8]{inputenc} % allow utf-8 input
\usepackage[T1]{fontenc}    % use 8-bit T1 fonts
\usepackage{hyperref}       % hyperlinks
\usepackage{url}            % simple URL typesetting
\usepackage{booktabs}       % professional-quality tables
\usepackage{amsfonts}       % blackboard math symbols
\usepackage{nicefrac}       % compact symbols for 1/2, etc.
\usepackage{microtype}      % microtypography
\usepackage{xcolor}         % colors

\usepackage{microtype}
\usepackage{graphicx}
\usepackage{subfigure}
\usepackage{booktabs} % for professional tables

\usepackage{amsthm}
\usepackage{latexsym}
\usepackage{amsmath}
\usepackage{cleveref}
\usepackage{amssymb}
\usepackage{epsfig}
\usepackage{fancybox}
\usepackage{pstricks}
\usepackage{url}
\usepackage{caption}
\usepackage{multirow}

\usepackage{fancyhdr}
\usepackage{color}
\usepackage{algorithm}
\usepackage{algorithmic}
\usepackage{wrapfig,lipsum}

\def\cD{{\cal D}}
\def\cL{{\cal L}}
\def\cN{{\cal N}}

\def\cX{{\cal X}}

\def\cS{{\cal S}}

\def\cC{{\cal C}}
\def\cU{{\cal U}}

\def\cW{{\cal W}}
\def\cO{{\cal O}}

\def\qed{\space$\square$ \par \vspace{.15in}}

\def\hat{\widehat}

\newcommand{\bz}{{\bf z}}

\newcommand{\bX}{{\bf X}}
\newcommand{\bx}{{\bf x}}

\newcommand{\bc}{\begin{center}}
	\newcommand{\ec}{\end{center}}
\newcommand{\be}{\begin{equation}}
	\newcommand{\ee}{\end{equation}}
\newcommand{\ba}{\begin{array}}
	\newcommand{\ea}{\end{array}}
\newcommand{\bean}{\begin{eqnarray*}}
	\newcommand{\eean}{\end{eqnarray*}}
\newcommand{\bea}{\begin{eqnarray}}
	\newcommand{\eea}{\end{eqnarray}}

\newtheorem{lemma}{\bf Lemma}

\newcommand{\ben}{\begin{enumerate}}
	\newcommand{\een}{\end{enumerate}}
\newcommand{\bed}{\begin{itemize}}
	\newcommand{\eed}{\end{itemize}}

\title{INN: A Method Identifying Clean-annotated Samples via Consistency Effect in Deep Neural Networks}

\author{  Dongha Kim  \\
		  Department of Statistics \\
		  Sungsin Women's University\\
		  \texttt{dongha0718@sungshin.ac.kr} \\
  \And
  Yongchan Choi  \\
  Department of Statistics \\
  Seoul National University\\
  \texttt{pminer32@gmail.com} \\
  \And
  Kunwoong Kim  \\
  Department of Statistics \\
  Seoul National University\\
  \texttt{kwkim.online@gmail.com} \\
  \And
  Yongdai Kim  \\
  Department of Statistics and \\ 
  Graduate School of Data Science \\
  Seoul National University\\
  \texttt{ydkim0903@gmail.com} \\
}

\begin{document}

\maketitle

\begin{abstract}
In many classification problems, collecting massive clean-annotated data is not easy,
and thus a lot of researches have been done to handle data with noisy labels.
Most recent state-of-art solutions for noisy label problems are built on the small-loss strategy which exploits the \textit{memorization effect}. While it is a powerful tool, the memorization effect has several drawbacks.
The performances are sensitive to the choice of a training epoch required for utilizing the memorization effect.
In addition, when the labels are heavily contaminated or imbalanced, the memorization effect
may not occur in which case the methods based on the small-loss strategy fail to identify clean labeled data.
We introduce a new method called \textit{INN} (Integration with the Nearest Neighborhoods) to refine clean labeled data from training data with noisy labels.
The proposed method is based on a new discovery that a prediction pattern at \textit{neighbor regions} of clean labeled data is consistently different from that of noisy labeled data regardless of training epochs.
The INN method requires more computation but is much stable and powerful than the small-loss strategy.  
By carrying out various experiments, we demonstrate that the INN method resolves the shortcomings in the memorization effect successfully and thus is helpful to construct more accurate deep prediction models with training data with noisy labels.
\end{abstract}

\section{Introduction}
\label{sec:introduction}

Learning deep neural networks (DNNs) has achieved impressive successes in many research fields but has suffered from collecting massive clean-annotated training samples such as ImageNet \cite{deng2009imagenet} and MS-COCO \cite{lin2014microsoft}. 
Since annotating procedures are usually done manually by human experts, it is expensive and  time-consuming to get large clean labeled data, which prevents deep learning models from being trained successfully. 
On the other hand, it is feasible to access numerous data through internet search engines \cite{fergus2010learning,schroff2010harvesting,xiao2015learning,krause2016unreasonable} or hashtags, whose labels are easy to collect but relatively inaccurate.
Thus it becomes to get a spotlight to exploit data sets with corrupted labels instead of clean ones to solve classification tasks with DNNs, which is called the \textit{noisy label problem}. 

There have been many kinds of literature dealing with noisy labeled data, and a majority of methods exploited so-called the \textit{memorization effect}, which is
a special characteristic of DNNs that DNNs memorize data eventually (i.e. perfectly classify training data) but memorize clean labeled samples earlier and noisy samples later \cite{arpit2017closer,jiang2018mentornet}. 
Hence, we can identify clean data from the given training data contaminated with noisy labels by choosing samples with small loss values. 
Due to its simplicity and superiority, many follow-up studies have been proposed based on the small-loss strategy and achieved great success (\cite{song2020learning} and references therein).

%%%%%%%%%%%%%%%%%%%%%%%%%%%%%%%%%%%%%%%
%% 210522- 일단 그림 없애자...나중에 다시 추가할 수도 있음.
%%%%%%%%%%%%%%%%%%%%%%%%%%%%%%%%%%%%%%%
%\begin{figure}[t]
%	\vskip 0.2in
%	\begin{center}
%		\centerline{ 
%			\includegraphics[width=0.4\textwidth]{./fig/comp_opt_me_epoch.png}}
%		\caption{The area under the ROC curve (AUC) values of clean/noisy classification of training data sets based on loss values for each training epoch. The best training epochs are highlighted with red dotted line. ({\bf Green}) AUC values of WideResNet28-2 (WRN) trained with the SGD optimizer of learning rate 2e-2 on noisy CIFAR100 and ({\bf Orange}) AUC values of PreActResNet18 (PRN) trained with the Adam optimizer of learning rate 2e-3 on noisy CIFAR10. 
%		}
%		\label{fig:opt_ep_comp}
%	\end{center}
%	\vskip -0.2in
%\end{figure}

\begin{wrapfigure}{r}{0.5\linewidth}
\centering
\includegraphics[width=0.49\textwidth]{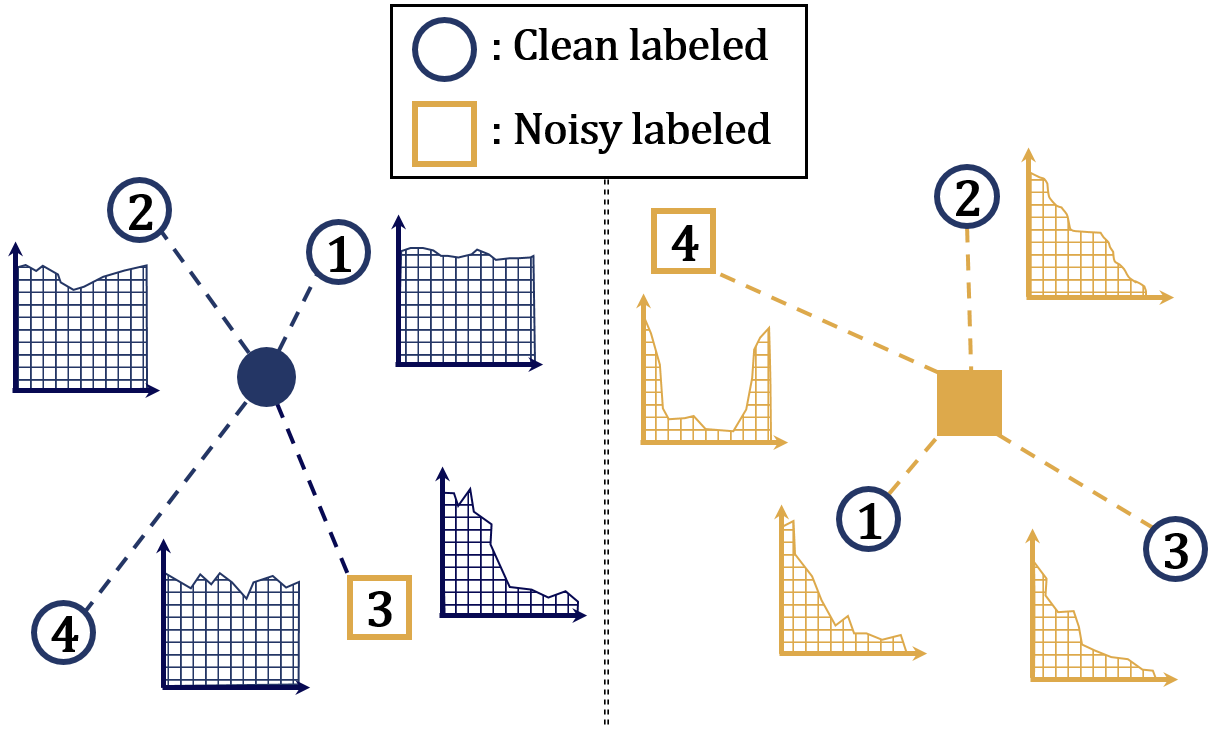}
\caption{An illustration of the INN method. Circle and square are inputs with clean label and noisy label, respectively. 
Numbered dots are the nearest inputs. 
Each graph presents the value of a given prediction model along the dashed line. 
The INN method takes an average of the areas under each of the graphs. 
}
\vspace{-.4cm}
\label{fig:inn_illustration}
\end{wrapfigure}

But the small-loss strategy has several weaknesses. 
First, during the training phase, it is difficult to know a training epoch (or iteration) where the discrepancy of loss values between clean data and noisy data is large since it heavily depends on various factors including data set, model architecture, optimizer type and even learning schedule. 
Second, it becomes hard to identify clean-annotated samples from training data via the small-loss strategy when the training labels are heavily polluted. 
Besides, the memorization effect may not appear when we analyze the data with imbalanced label distribution. 
%To be more specific, we found that the memorization effect could fail to refine the clean samples belonging to the minor class.
As we can obtain imbalanced data frequently in many real-world domains, this shortcoming can be an obstacle for the small-loss strategy applied in many industry fields.  
%See Figure \ref{fig:opt_ep_comp} for detailed illustrations. 

%As we can frequently observe data with imbalanced classes in many real-world applications such as ??????, this issue prevents the small-loss-based algorithms from being utilized in many domains. 
%\cite{Liu2020EarlyLearningRP,pmlr-v97-song19b}. 
%And if a model memorizes all the training samples, we cannot distinguish clean and noisy samples anymore via their loss values.  
%So, it would be problematic for the {memorization effect} based methods to be adopted to many application fields in practice.

To tackle these issues about the memorization effect, we develop a novel and powerful method called \textit{INN} ({\bf I}ntegration with the {\bf N}earest {\bf N}eighbors). %, which outputs scores of cleanness of samples. 
We start with a new and interesting observation that the output values of a trained DNN at \textit{neighbor regions} of labeled and noisy samples are consistently much different regardless of training epochs. 
%Our study starts off to closely investigate the behavior of a trained DNN at training sample's neighbor regions. 
%We start with a new and interesting observation that the output values of a trained DNN at neighbor regions 
We call this phenomenon the \textit{consistency effect}.
%\item Our method is simple, powerful and theoretically grounded. 
%\item We develop a new method, called \textit{INN} ({\bf I}ntegration with the {\bf N}earest {\bf N}eighbors), which outputs scores of cleanness of samples. 
%The CE is observed simply because samples locating at a neighborhood of a given noisy sample have different labels from the label of the noisy sample.
Motivated by the consistency effect, 
%the INN takes advantage of the output values of a neighbor region of a given sample by integration operation. 
the INN method takes averages 
of the output values of neighbor regions of a given sample and decides it as noisy
if the average is small. See Figure \ref{fig:inn_illustration} for an illustration of the INN method.
%For a given prediction model and a given sample, \textit{INN} conducts integration operation of the prediction probability over intervals between the input and its nearest train inputs in order to utilize advantage of consistency effect.
%\textit{INN} exploits the consistency effect as well as the memorization effect, which facilitates to refine clean labeled data from corrupted ones better than those using the memorization effect only.

%\begin{figure}[t]
%	\vskip 0.2in
%	\begin{center}
%		\centerline{ 
%			\includegraphics[width=0.4\textwidth]{./fig/inn_illustration3.png}}
%		\caption{An illustration of the \textit{INN} method. Circle and square are inputs with clean label and noisy label respectively. 
%			Numbered dots are the nearest inputs. 
%			Each graph presents the value of a given prediction model along the dashed line. %
%			The \textit{INN} method takes an average of the areas under each of the graphs. 
%		}
%		\label{fig:inn_illustration}
%	\end{center}
%	\vskip -0.2in
%\end{figure}

In fact, the INN requires more computation than the small-loss method. 
Still, this additional expense deserves to pay since the INN successfully overcomes the small-loss method's limitations. 
%The INN can distinguish clean samples from noisy ones stably regardless of the choice of a training epoch in the training phase.  
%One of the  most appealing features of the \textit{INN} is that 
%the discrepancy between clean and noisy samples
%is not sensitive to the choice of a training epoch in the
%training phase. 
The INN works well even when the training labels are heavily contaminated or has imbalanced distribution, while the small-loss method is in trouble for the situations. 
The stability and superiority make the INN easily applicable to various supervised learning tasks without much effort. 
%This insensitivity  makes the \textit{INN} easily applicable to various supervised learning problems.

We can also combine the INN with an existing noisy-label-problem-solving learning method based on the small-loss strategy (e.g. DivideMix \cite{li2020dividemix}) to construct deep networks of high accuracy. 
We replace the parts where the memorization effect and loss information are used with the consistency effect and the INN information. 
%We propose a simple modification by replacing the training data's cleanness decision part using their loss scores with the INN scores. 
We show that these modifications enhance prediction performances much, especially when training labels have many noises or imbalanced distribution. 

This paper is organized as follows. 
%In Section \ref{sec:related_works}, we provide brief reviews  for related studies dealing with noisy labels, and detailed descriptions of two proposed methods, the \textit{INN} and the \textit{INN-DivideMix}, are given in Section \ref{sec:inn} and \ref{sec:app_inn} respectively.
In Section \ref{sec:related_works}, we provide brief reviews  for related studies dealing with noisy labels, and detailed descriptions of the INN are given in Section \ref{sec:inn}.
Various experimental analyses including performance test and ablation study are given in Section \ref{sec:experiment} and final concluding remarks follow in Section \ref{sec:conclusions}. The key contributions of this work are as follows.
\bed 
\item We find a new observation called the consistency effect, that the output values of a trained DNN at neighbor regions of labeled and noisy samples are consistently much different regardless of training epochs. 
\item Built on the consistency effect, we propose a method called the INN to identify clean annotated data from a given training data.
\item We empirically demonstrate that the INN can separate clean and noisy samples accurately and stably even under the heavy label corruption and imbalanced label distribution, and also helpful to construct superior prediction models.
%\item We provide a new learning framework called \textit{INN-DivideMix} which trains prediction models by combining the \textit{INN} and DivideMix.
%%%%%%%%%%%%%%%%%%%%%%%%%%%%%%%%%%%%%%%%%%%%%%
%\item We empirically demonstrate superiority and stability of the \textit{INN} and \textit{INN-DivideMix} by analyzing several well-known benchmark data sets. 
\eed

\section{Related works}
\label{sec:related_works}

The  \textit{noisy label problem} has been studied for several decades \cite{angluin1988learning,zhu2004class,nettleton2010study}. 
The core issue to solve the noisy label problem with DNNs is that DNNs easily over-fit all training samples, including noisy labeled ones, because of too large complexities resulting in inferior generalization performances. 
Here we review some related studies for efficient algorithms to train robust classifiers in noisy annotations based on the key concepts called the \textit{loss correction} and the \textit{memorization effect}.  
And we also describe several approaches exploiting the information of a target sample's neighborhoods as the INN does. 
%The strategies to overcome this issue are naively categorized by the two key concepts: (i) \textit{loss correction} and the (ii) \textit{memorization effect} (ME).

%\paragraph{Loss correction} 
Loss correction based algorithms have a goal to improve the generalization error by modifying objective functions \cite{patrini2017making,zhang2018generalized}. The noise adaptive layer-based algorithm \cite{goldberger2016training} added additional noisy channels that estimate the correct labels. The iterative noisy label detection \cite{wang2018iterative,thulasidasan2019combating} used the weighted softmax loss function where the weights are updated iteratively based on the feature maps of the current DNN model. Some algorithms to estimate ground-truth labels directly have been developed \cite{tanaka2018joint,yi2019probabilistic}. The meta-learning algorithm was also applied to resolve the noisy label problem \cite{li2019learning}.
There was an attempt to propose a new loss function more robust than standard loss functions \cite{Lyu2020Curriculum}. 

%\paragraph{Memorization effect} 
Approaches based on the memorization effect focused on the gap between the output values of clean labeled and noisy labeled samples during an early stage of the training phase. The decouple method \cite{NIPS2017_58d4d1e7} proposed a meta-algorithm called decoupling which decides when to update. D2L \cite{pmlr-v80-ma18d} distinguished clean labeled data from noisy ones by employing a local dimensionality measure and ELR \cite{Liu2020EarlyLearningRP} found the faster gradient vanishings of clean labeled samples at the early learning stage. 
There were several algorithms to train noisy-robust prediction models by using only a subset of the training data based on their loss or prediction values. \cite{10.5555/3327757.3327944,pmlr-v97-yu19b,pmlr-v97-shen19e,chen2019understanding,pmlr-v97-song19b,Nguyen2020SELF:}. Some studies fitted a two-component mixture model to a per-sample loss distribution \cite{arazo2019unsupervised,li2020dividemix}. 

%\paragraph{Utilizing neighbor's information}
%\kw{ 
Some works tried to utilize the information of neighborhoods to filter out noisy labeled data, similar to the INN's idea.
%There are some recent works related to INN in that share the information of target input's  neighborhoods. 
%techniques in a broader view, while the core idea of them and INN are significantly different. 
The distance-weighted $k$-NN \cite{weightedknn} was an initial work that considers the nearest samples with their distance-based weights. 
Deep $k$-NN \cite{deepknn} proposed a filtering strategy based on the label information of nearest neighbors, and MentorMix \cite{mentormix} applied MixUp \cite{zhang2017mixup} to MentorNet \cite{jiang2018mentornet} to consider linear combinations of two inputs. 
%, whereas it is still based on the ME. 
We will discuss about the difference between the INN and the methods of exploiting neighbor's information in Section \ref{sec:measure}. 
%more precise differences between them and INN after introducing INN in Section \ref{sec:measure}.
%}

\section{Integration with the nearest neighbors}
\label{sec:inn}
\subsection{Notations and definitions}
\label{sec:notations}

For a given input vector $\bx\in\mathbb{R}^d$, let $y,y^*\in[K]$ be its observable and ground-truth labels, respectively, where $[K]=\{1,\ldots,K\}$. 
Of course, $y$ might be different from $y^*$. 
We say that the sample $(\bx,y)$ is cleanly labeled if $y=y^*$ and noisily labeled if $y\neq y^*$.
Let $\cD^{\text{tr}}=\left\{ (\bx_i,y_i) \right\}_{i=1}^n$ be a training data set with $n$ samples. %, where $\bx_i\in\mathbb{R}^d$ is the $i$-th input sample and $y_i\in[K]$ is the corresponding label. 
Define $\cC^{\text{tr}}=\{(\bx,y)\in\cD^{\text{tr}}:y=y^*\}$ and $\cN^{\text{tr}}=\{(\bx,y)\in\cD^{\text{tr}}:y\neq y^*\}$. 
%And we also denote $\cD_{\bx}$ as the set of train inputs.
Our goal is to identify the clean labeled subset $\cC^{\text{tr}}$ from $\cD^{\text{tr}}$ accurately.

Let $f(\bx;\theta):\mathbb{R}^d\to\mathbb{R}^K$ (abbr. $f(\bx)$) be a discriminative DNN parametrized by $\theta$ which maps an input $\bx$ to a $K$-dimensional conditional probability vector with the softmax layer. 
Also let $f_k(\bx)$ be the $k$-th component of $f(\bx)$, that is, we can represent $f(\bx)$ as $(f_1(\bx),\ldots,f_K(\bx))$. \
%And for a given prediction model $f(\bx;\theta)$, denote the feature vector of the penultimate layer as $h(\bx;\theta)$. 

%Finally we define $\hat{\theta}_t$ an estimated parameter vector with some loss function at the train epoch $t$. 

\subsection{Consistency effect}
\label{sec:motivation}

\begin{figure}[t]
   \begin{minipage}{0.33\textwidth}
\bc
\includegraphics[width=\textwidth]{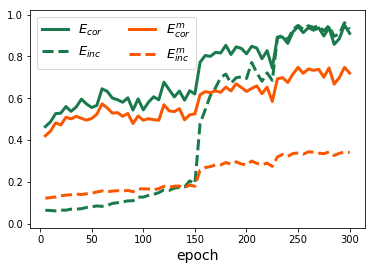}
\ec
\caption{Comparison of $E_{\text{cor}}$, $E_{\text{inc}}$, ${E}_{\text{cor}}^{\text{m}}$ and ${E}_{\text{inc}}^{\text{m}}$ for various training epochs. 
We use the 30\% symmetrically noisy CIFAR10 as $\cD^{\text{tr}}$. 
}
\label{fig:motivation}
   \end{minipage}\hfill
   \begin{minipage}{0.65\textwidth}
\bc
\vspace{-.1cm}
\includegraphics[width=0.49\textwidth]{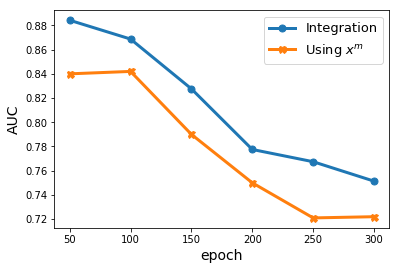}
\includegraphics[width=0.49\textwidth]{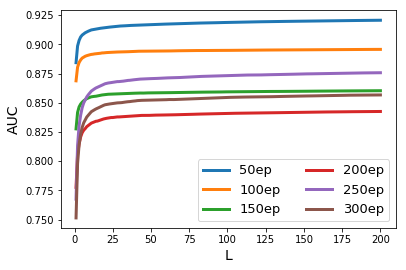}
\ec
\caption{Comparison of AUC values for clean/noisy sample classification of the INN by varying some factors. 
We consider about ({\bf Left}) how to utilize the prediction information of neighbor regions and ({\bf Right}) the number of the nearest neighborhoods. 
We use the 40\% asymmetrically noisy CIFAR10 as $\cD^{\text{tr}}$
}
\label{fig:inn_factors}
   \end{minipage}
\end{figure}

Before we start, we explain the main motivation of our method. 
%For a given training sample $(\bx,y)\in\cD^{\text{tr}}$, most studies inspired by the ME try to find noisy sample by using the values of $f_y(\bx)$
%at each sample. In contrast, we conduct investigation for the values of $f_y$ at a \textit{periphery} of $\bx$ of each sample. 
For a given training sample $(\bx,y)\in\cD^{\text{tr}}$, we define $\bx^{\text{m}}=(\bx+\tilde{\bx})/2$, where $\tilde{\bx}$ is the nearest neighbor training input of $\bx$ on the feature space $h(\cdot;\hat{\eta})$ (i.e. $h(\tilde{\bx};\hat{\eta})$ is most close to $h(\bx;\hat{\eta})$) and $h(\cdot;\hat{\eta})$ is the output of the penultimate layer of a pre-trained prediction model on $\cD^{\text{tr}}$. 
Then, we can regard that $\bx^{\text{m}}$ locates in the neighbor region of $\bx$. 
We investigate how the prediction values of the training inputs and their neighbors behave differently by the label cleanness.
%We analyze how the values $f_y(\bx)$ and $f_y(\bx^{\text{m}})$ behave with respect to the label cleanness of $\bx$.  
We also estimate a prediction model $f(\bx;\hat{\theta})$ by minimizing the standard cross-entropy based on $\cD^{\text{tr}}$.  
At each training epoch, we calculate the four expectations defined as 
\bean
\begin{array}{ll}
E_{\text{cor}}=\mathbb{E}_{(\bx,y)\sim \cC^{\text{tr}}}\left[f_y(\bx)\right], &  
E_{\text{inc}}=\mathbb{E}_{(\bx,y)\sim \cN^{\text{tr}}}\left[f_y(\bx)\right]\\
{E}_{\text{cor}}^{\text{m}}=\mathbb{E}_{(\bx,y)\sim \cC^{\text{tr}}}\left[f_y(\bx^{\text{m}})\right], &
{E}_{\text{inc}}^{\text{m}}=\mathbb{E}_{(\bx,y)\sim \cN^{\text{tr}}}\left[f_y(\bx^{\text{m}})\right].
\end{array}
\eean
Two values $E_{\text{cor}}$ and $E_{\text{inc}}$ are the expectations of clean and noisy data's predictions, respectively, and ${E}_{\text{cor}}^{\text{m}}$ and ${E}_{\text{inc}}^{\text{m}}$ are the expectations of neighbor region's prediction values over the clean and noisy labeled data, respectively.
From Figure \ref{fig:motivation}, we can see a typical phenomenon related to the memorization effect:
$E_{\text{cor}}$ and $E_{\text{inc}}$ are much different at early epochs but
the difference diminishes as the training epoch proceeds.
That means that it becomes hard to discriminate noisy data from clean ones by comparing $f_y(\bx)$ values at each sample at later stages of the training phase. 
But it is difficult to decide how many epochs are necessary for amply utilizing the memorization effect.

On the other hand, the difference between ${E}_{\text{cor}}^{\text{m}}$ and ${E}_{\text{inc}}^{\text{m}}$ is clearly significant regardless of training epochs. 
That is, the prediction values of neighbor regions for each sample are informative to separate clean and noisy data even when the number of training epoch is large. 
We call this new observation the \textit{consistency effect}. 
This consistent discrepancy occurs by two reasons. 
%First, samples locating at a neighborhood of a given noisy sample have different labels from the label of the noisy sample.
First, when $(\bx,y)$ is a noisy labeled sample, the label of an input $\bx$, i.e. $y$, and the label of its neighborhood training sample $\tilde{\bx}$ denoted by $\tilde{y}$ tend not to coincide, which yields the small prediction value $f_y(\bx^{\text{m}})$. 
And even if $y$ and $\tilde{y}$ are equal, 
%is that for a given sample $(\bx,y)\in\cD^{\text{tr}},$ if the label $y$ is noisy, 
$\tilde{\bx}$ may not be the nearest neighbor on the input space (i.e. $\tilde{\bx}$ is not close to $\bx$).
Hence, there exists a region between $\bx$ and $\tilde{\bx}$ at which the value of $f_y(\cdot)$ becomes small. 
%, and it is the reason why ${E}_{\text{inc}}^{\text{m}}$ is smaller than ${E}_{\text{cor}}^{\text{m}}$.

%During the training procedure, the prediction values of neighbor regions of a given input are affected by the information of labels of inputs locating at a neighborhood of the input, and the majority labels of neighborhoods are cleanly labeled. 
%Thus, for a given sample $(\bx,y)$, the $y$-th prediction value at neighbor regions would be high if $y$ is clean and vice versa.
%That is why the \textit{consistency effect} occurs throughout whole training epochs.
%and it is the key observation which gives us a motivation to adopt the information of data \textit{near} the train data as well as the train data themselves to resolve the noisy label problem.

%\begin{figure}[t]
%	\vskip 0.2in
%	\begin{center}
%		\centerline{ 
%			\includegraphics[width=0.37\textwidth]{./fig/inn_motivation.png}}
%		\caption{Comparison of $E_{\text{cor}},E_{\text{inc}},{E}_{\text{cor}}^{\text{m}}$ and ${E}_{\text{inc}}^{\text{m}}$ for various training epochs. 
%			We use the 30\% symmetrically noisy CIFAR10 as the training data $\cD^{\text{tr}}$. 
%		}
%		\label{fig:motivation}
%	\end{center}
%	\vskip -0.2in
%\end{figure}

\subsection{{INN} method}
\label{sec:measure}

%\begin{figure*}[t]
%	\vskip 0.2in
%	\begin{center}
%		\centerline{
%			\includegraphics[width=0.32\textwidth]{./fig/integration_vs_simple_averaging.png}
%			\includegraphics[width=0.32\textwidth]{./fig/nnbhd_comp_cifar10.png}
%		}
%		\caption{Comparison of AUC values for clean/noisy sample classification of the INN by varying some factors. 
%		We consider ({\bf Left}) how to utilize the prediction information of neighbor regions and ({\bf Right}) the number of the nearest neighborhoods. 
%       }
%		\label{fig:inn_factors}
%	\end{center}
%	\vskip -0.2in
%\end{figure*}

In this section, we propose a new and novel method to identify clean labeled samples motivated by the consistency effect.
%In this section, we introduce our \textit{INN} method to refine data with noisy labels.
As being observed in Section \ref{sec:motivation}, it is important to take into account the prediction values at a neighbor region of each sample. 
Let $f(\cdot;\hat{\theta})$ be a prediction model trained with a loss function $l$ on $\cD^{\text{tr}}$ for $T^{\text{inn}}$ training epochs. 
In this study, we use the MixUp objective function \cite{zhang2017mixup} as the loss function $l$. 
For a given training sample $\bz=(\bx,y)\in\cD^{\text{tr}}$ and its neighborhood training input $\tilde{\bx}$, a naturally induced score to identify whether $y$ is clean or not would be  $I(\bx;\hat{\theta},\tilde{\bx}):=f_y(\bx^{\text{m}};\hat{\theta})$, where $\bx^{\text{m}}=(\bx+\tilde{\bx})/2$. 
From further experiments, we modify the score as follows. 
First, it is observed that the consistency effect occurs at many input vectors between $\bx$ and $\tilde{\bx}$ other than $\bx^{\text{m}}$. 
Thus, to exploit the consistency effect fully, we consider integrating the prediction function over the whole interval between $\bx$ and $\tilde{\bx}$ to have
\bean
I(\bx;\hat{\theta},\tilde{\bx})=\int_{0}^1 f_y\left(\alpha\bx+(1-\alpha)\tilde{\bx};\hat{\theta}\right)d\alpha.
\eean
Second, using multiple neighbor samples helps identify clean labeled data more accurately. 
%For a given prediction model $f(\cdot;\hat{\theta})$ trained with a loss function $l$ on $\cD^{\text{tr}}$ for $T^{\text{inn}}$ training epochs and a training sample $\bz=(\bx,y)\in\cD^{\text{tr}}$, let $\tilde{\cX}=\{\tilde{\bx}_1,\ldots,\tilde{\bx}_L\}$ be the set of $L$ nearest neighbor training inputs of $\bx$ on the feature space described in Section \ref{sec:motivation}. 
Based on these two arguments, we propose the INN score given as:
%Thus, the final score, called the INN score, is of the form:
%The \textit{INN} score of $\bz$ denoted by $I(\bz;\hat{\theta},\tilde{\cX})$ (abbr. $I(\bz)$) is defined as the mean of $L$ integration values given as
\bea
\label{eq:inn}
I(\bz;\hat{\theta},\tilde{\cX})=\frac{1}{L}\sum_{l=1}^L \int_0^1 f_y(\alpha\bx+(1-\alpha\tilde{\bx}_l);\hat{\theta})d\alpha,
\eea
%where the approximation is established by the trapezoidal rule and $\bx_{l,h}=\frac{H-h}{H}\bx+\frac{h}{H}\tilde{\bx}_l.$ 
where $\tilde{\cX}=\{\tilde{\bx}_1,\ldots,\tilde{\bx}_L\}$ is the set of $L$ nearest neighbor training inputs of $\bx$ on the feature space described in Section \ref{sec:motivation}. 
Figure \ref{fig:inn_factors} illustrates the effects of these two modifications. 
The integration in (\ref{eq:inn}) can be easily approximated by the trapezoidal rule as follows:
\bean
\label{eq:inn_apprx}
\frac{1}{L}\sum_{l=1}^L\sum_{h=1}^H \frac{1}{2H}\left(f_y(\bx_{l,h-1};\hat{\theta})+f_y(\bx_{l,h};\hat{\theta})\right),
\eean
where $\bx_{l,h}=\frac{H-h}{H}\bx+\frac{h}{H}\tilde{\bx}_l$ and $H$ is the number of trapezoids. 
In practice, we fix the value of $H$ and $L$ to 10. 
%We find that considering $L$ larger than one gives stable refinement result. 
%and We consider multiple integration values and take  larger than one due to the stability of the \textit{INN}.
The larger the score $I(\bz;\hat{\theta},\tilde{\cX})$ is, the more we could regard $\bz$ as being cleanly labeled. 
Hereafter, we will abbreviate $I(\bz;\hat{\theta},\tilde{\cX})$ to $I(\bz)$. 
Even after large training epochs, since there still remains the consistency effect in the prediction model, the INN method separates clean labeled data from noisy ones well. 
The following simple lemma supports the validity of our method. 

%\renewcommand{\algorithmiccomment}[1]{#1}
%\begin{algorithm}[tb]
%	\caption{INN}
%	\label{alg:inn}
%	\begin{algorithmic}[1]
%		\INPUT Training data $\cD^{\text{tr}}=\{(\bx_i,y_i)\}_{i=1}^n$, a prediction model $f(\cdot;{\theta})$, a pre-trained feature model $h(\cdot;\hat{\eta}^{\text{pt}})$, the number of neighborhoods $L$, a training epoch $T^{\text{inn}}$, an optimizer $\cO$
		
%		\STATE $D(\cdot,\cdot)\leftarrow L_2(h(\cdot;\hat{\eta}^{\text{pt}}),h(\cdot;\hat{\eta}^{\text{pt}}))$ \hfill \algorithmiccomment //dissimilarity measure
%		\FOR {$t=1$ {\bfseries to} $T^{\text{inn}}$}
%		\STATE $\hat{\theta} \leftarrow MU(f(\cdot;\theta), \cD^{\text{tr}},\cO)$  \hfill \algorithmiccomment //training
%		\ENDFOR
		
%		\STATE $\cS\leftarrow\emptyset$    \hfill \algorithmiccomment //INN score set
%		\FOR{$i=1$ {\bfseries to} $n$}
%		\STATE $\tilde{\cX}_i=\{\tilde{\bx}_{i1},\ldots,\tilde{\bx}_{iL}\}\leftarrow neighbor(\bx_i,D)$ \\
%		\hfill \algorithmiccomment //$L$ nearest neighborhoods of $\bx_i$
%		\STATE $\bz_i\leftarrow (\bx_i,y_i)$
%		\STATE $s_i\leftarrow I(\bz_i;\hat{\theta},\tilde{\cX}_i)$ 
%		\hfill \algorithmiccomment //INN score of $\bz_i$
%		\STATE $\cS\leftarrow \text{append}(\cS,s_i)$ \hfill \algorithmiccomment //append $s_i$ to $\cS$
%		\ENDFOR
%		\OUTPUT $\cS=\{s_i\}_{i=1}^n$
%	\end{algorithmic}
%\end{algorithm}

\begin{lemma}
\label{lem:1}
Let $f:\mathbb{R}^d\to\mathbb{R}^K$ be a prediction model which perfectly over-fits the MixUp loss function 
\bean
\mathbb{E}_{(\bx_1,y_1),(\bx_2,y_2)\sim\cD^{tr}}\mathbb{E}_{\lambda\sim B(\alpha,\alpha)}\left[\text{CE}\left(\text{mix}_{\lambda}(y_1,y_2), f(\text{mix}_{\lambda}(\bx_1,\bx_2) \right)\right],
\eean
where CE is the cross-entropy loss function, $\text{mix}_\lambda(a,b)=\lambda a+(1-\lambda)b$, and $B(\alpha,\alpha)$ is the Beta distribution with a hyperparameter $\alpha>0$. 
Also let assume that for each training input $\bx$, its nearest neighbor set  $\cX=\{\tilde{\bx}_1,\ldots,\tilde{\bx}_L\}$ satisfies $\text{argmax}_{k\in[K]} \sum_{l=1}^L I(\tilde{y}_l=k)=y^*$, where $\tilde{y}_l$ is the observed label of $\tilde{\bx}_l$ and $y^*$ is the ground-truth label of $\bx$. Then, the following inequality holds:
\bean
\min_{\bz\in\cC^{tr}}I(\bz)> \max_{\bz\in\cN^{tr}}I(\bz).
\eean
\end{lemma}

  \begin{wrapfigure}{R}{0.6\textwidth}
    \begin{minipage}{0.6\textwidth}
    \vspace{-.5cm}
\begin{algorithm}[H]
	\caption{INN (In practice, we fix $L$ to 10.)}
	\label{alg:inn}
	\begin{algorithmic}[1]
		\INPUT Training data $\cD^{\text{tr}}=\{(\bx_i,y_i)\}_{i=1}^n$, a prediction model $f(\cdot;{\theta})$, a pre-trained feature model $h(\cdot;\hat{\eta})$, the number of neighborhoods $L$, a training epoch $T^{\text{inn}}$, an optimizer $\cO$
		
		%\FOR {$t=1$ {\bfseries to} $T^{\text{pt}}$}
		%\STATE $\hat{\eta}^{\text{pt}}\leftarrow CE(h(\cdot;\eta), \cD^{\text{tr}},\cO)$ \hfill \algorithmiccomment //pre-training
		%\ENDFOR
		\STATE $D(\cdot,\cdot)\leftarrow L_2(h(\cdot;\hat{\eta}),h(\cdot;\hat{\eta}))$ \hfill  //dissimilarity measure
		\FOR {$t=1$ {\bfseries to} $T^{\text{inn}}$}
		\STATE $\hat{\theta} \leftarrow MixUp(f(\cdot;\theta), \cD^{\text{tr}},\cO)$  \hfill  //train $f$ using MixUp
		\ENDFOR
		
		%\STATE {\bfseries Input:} 
		%\STATE {\bfseries (Pre-train step)}
		%\STATE $\hat{\theta}^{\text{pt}}=MixUp(\cD^{\text{tr}};T_1)$   
		%\STATE $\cH=\{h(\bx_i;\hat{\theta}^{\text{pt}})\}_{i=1}^n$	
		%\STATE {\bfseries (INN step)}
		\STATE $\cS\leftarrow\emptyset$    \hfill  //Define INN score set
		\FOR{$i=1$ {\bfseries to} $n$}
		%\FOR{$\bz=(\bx,y)$ {\bfseries in} $\cD^{\text{tr}}$}
		\STATE $\tilde{\cX}_i=\{\tilde{\bx}_{i1},\ldots,\tilde{\bx}_{iL}\}\leftarrow neighbor(\bx_i,D)$ \\
		\hfill  //$L$ nearest neighborhoods of $\bx_i$
		%\STATE Search $L$ neighborhoods of the input $\bx_i$, $\tilde{\bx}_{i1},\ldots,\tilde{\bx}_{iL}$ by use of $D$.
		\STATE $\bz_i\leftarrow (\bx_i,y_i)$
		\STATE $s_i\leftarrow I(\bz_i;\hat{\theta},\tilde{\cX}_i)$ 
		\hfill  //INN score of $\bz_i$
		\STATE $\cS\leftarrow \text{append}(\cS,s_i)$ \hfill  //append $s_i$ to $\cS$
		\ENDFOR
		\OUTPUT $\cS=\{s_i\}_{i=1}^n$
	\end{algorithmic}
\end{algorithm}
    \end{minipage}
  \end{wrapfigure}
  
The proof is in the supplementary materials.  
Lemma \ref{lem:1} means that if we have a prediction model trained with the MixUp and good nearest neighborhood sets, then the INN separates $\cC^{tr}$ perfectly from $\cN^{tr}$. 

As mentioned in Section \ref{sec:related_works}, there are several approaches similar to the INN that utilize the nearest neighborhoods' information to filter out noisy data \cite{wang2018iterative,li2019learning,deepknn,mentormix}. 
They mainly take advantage of the labels of the neighborhoods. 
When the training labels are polluted heavily, most of the nearest samples also become noisy, thus relying only on the label information might lead to bad results. 
On the other hand, the INN focuses on the regions between the inputs and their neighbor training samples.  So, we expect that the INN would be robust to highly noisy data. 
%\kw{However, they are probably not robust when the noise rate is high: if the nearest samples are mostly noisy, then the corresponding (clean) score of the target input might goes lower since the label information are wrong. Furthermore, MentorMix \cite{mentormix} and INN both consider not only the nearest samples but the linear regions between the target input and its neighbors, however INN does not apply the ME while MentorMix does.}

The algorithm of the \textit{INN} method is summarized in Algorithm \ref{alg:inn}.

\section{Experimental analysis}
\label{sec:experiment}

In this section, we empirically show the superiority of the INN in terms of three aspects. 
First, the INN is not sensitive to the choice of the training epochs and provides consistent performances. 
Second, the INN is significantly better than the small-loss strategy when many polluted labels are in the training data. 
Finally, in a situation where the training labels are imbalanced, the small-loss strategy may not work, while our method still succeeds in finding clean labeled data. 
And we also provide a combination of the INN and an existing small-loss-based learning framework to construct better deep prediction networks.
Some additional ablation studies follow after then. 

\subsection{Experimental settings}
\paragraph{Data sets}
%\subsection{Data sets}
We carry out extensive experiments including performance tests and ablation studies by analyzing three data sets, CIFAR10\&100 \cite{krizhevsky2009learning} and Clothing1M \cite{xiao2015learning}. 
Both CIFAR10 and CIFAR100 consist of 50K training data and 10K test data with an input size of $3\times 32\times 32$ all of which
are cleanly labeled. 
Clothing1M is a large-scale data set with real-world noisy labels containing 1M training data collected from online shopping websites. 
We use the subset of the Clothing1M data set whose ground-truth labels are known. 
The subset consists of 48K samples with a noisy level of 20\% roughly.

As for imposing noise labels to CIFAR10 and CIFAR100, we consider symmetric and asymmetric settings as other studies did  %\cite{tanaka2018joint,zhang2018generalized,li2019learning,thulasidasan2019combating,yi2019probabilistic}. 
\cite{zhang2018generalized,yi2019probabilistic}. 
In the symmetric noise setting, for each sample in the training data set, its label is contaminated with a probability $r$ to a random label generated from the uniform distribution on $1$ to $K$ ($K=10$ for CIFAR10 and $K=100$ for CIFAR100). 
In the asymmetric noise setting for CIFAR10, with a probability $r$, a noisy label is generated by one of the following mappings: \emph{truck$\to$automobile, bird$\to$airplane, deer$\to$horse} and \emph{cat$\leftrightarrow$dog}.
For CIFAR100, labels are asymmetrically contaminated by flipping a given label to the next label with a probability $r$ according to the transition chain: \emph{class1$\to$class2$\to\cdots\to$class100$\to$class1}.

\paragraph{Architectures and implementation details}
%\subsection{Implementation details}

We need two models $f$ and $h$, and we use the same architectures for them in all experiments. For CIFAR10\&100 we utilize PreActResNet18 (PRN, \cite{he2016identity}) with randomly initialized weights, and for Clothing1M we use ResNet50 (RN, \cite{szegedy2015going}) with pre-trained weights by ImageNet.
%For CIFAR10 and CIFAR100, we utilize PreActResNet18 (PRN, \cite{he2016identity}), %, and optimize them with the SGD with a momentum 0.9. % or the Adam \cite{kingma2014adam}. 
%and for Clothing1M, ResNet50 (RN, \cite{szegedy2015going}) with initialized weights pre-trained by ImageNet is used. % and trained with SGD with a momentum 0.9. 
We train all the deep networks using the SGD algorithm with a momentum of 0.9 and the mini-batch size of 128, set the initial learning rate as 0.02, and reduce it by a factor of 5 when the half and three-fourths of the learning procedure proceed, respectively. 
%We set the initial learning rate as 2e-2 for the SGD and 2e-3 for the Adam respectively, and reduce them by a factor of 10 after 150 epochs.
%{\bf Unless otherwise stated, we fix the total training epochs,$T^{\text{tot}}$, and the mini-batch size to 300 and 128 respectively.} 
%Finally, we refer \cite{li2020dividemix} for the hyperparameter settings, e.g. the tuning parameter in the loss function, of the SSL phase of the DivideMix. 
All the results of ours in the following experiments are the averaged values of three trials executed from random initial weights and mini-batch arrangements. 

%but train only $f$ (i.e. let $T^{\text{pt}}=1$ in the \textit{INN} algorithm).
%estimate parameters with the SGD algorithm with a momentum 0.9 and a batch size 128. 
\begin{figure*}[t]
\vskip 0.2in
\begin{center}
\centerline{
\includegraphics[width=0.245\textwidth]{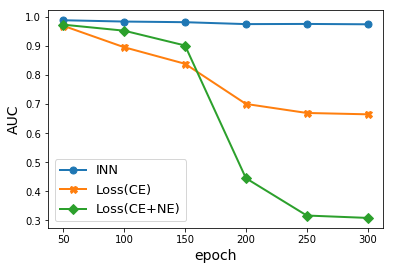}
\includegraphics[width=0.245\textwidth]{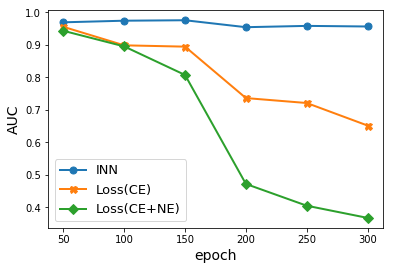}
\includegraphics[width=0.245\textwidth]{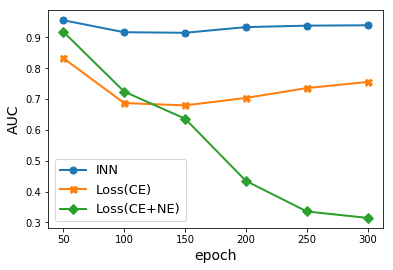}
\includegraphics[width=0.245\textwidth]{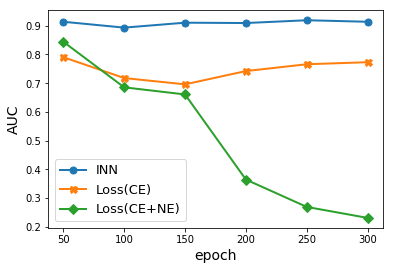}
}
\centerline{
\includegraphics[width=0.245\textwidth]{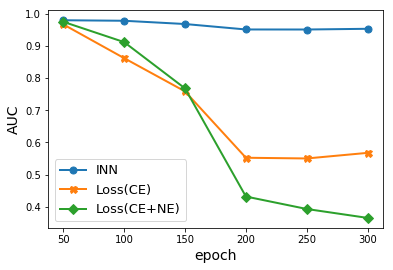}
\includegraphics[width=0.245\textwidth]{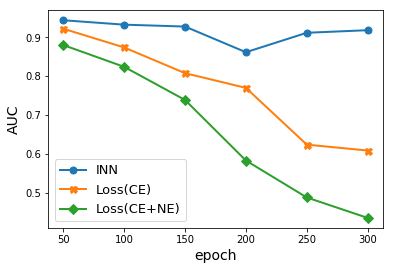}
\includegraphics[width=0.245\textwidth]{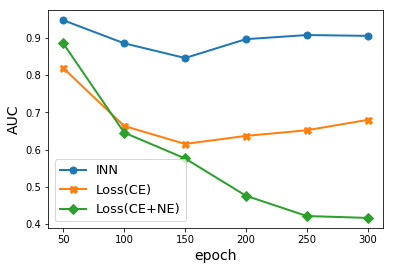}
\includegraphics[width=0.245\textwidth]{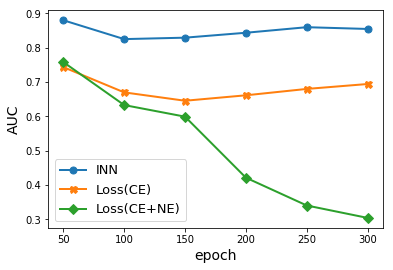}
}
\caption{Comparison of AUC values for clean/noisy sample classification between the \textit{INN} and the small-loss methods with two losses. 
The plot positioned at the $i$-th row from the top and the $j$-th column from the left is the result of the Case$i$-$j.$ 
}
\label{fig:cifar10_100_auc}
\end{center}
\vskip -0.2in
\end{figure*}

\subsection{Stability test of the INN}
%\subsection{Performance test of \textit{INN}}
\label{sec:exp-stability}

%\paragraph{CIFAR10\&100}
In this section, we show the stability and superiority of the INN for identifying clean labeled samples from training data. 
%We analyze CIFAR10\&100 and Clothing1M data sets. 
For CIFAR10\&100, we consider eight cases (Case1-1 to Case4-2) by varying noise rates and noise types. % given as:
\bed
\item CIFAR10 with (Case1-1) $r=0.1$ and (Case1-2) $r=0.3$ symmetrically noisy labels
\item CIFAR10 with (Case2-1) $r=0.1$ and (Case2-2) $r=0.3$ asymmetrically noisy labels
\item CIFAR100 with (Case3-1) $r=0.2$ and (Case3-2) $r=0.4$ symmetrically noisy labels
\item CIFAR100 with (Case4-1) $r=0.1$ and (Case4-2) $r=0.2$ asymmetrically noisy labels
\eed

\begin{figure}
\vspace{-.4cm}
\centering
\includegraphics[width=0.37\textwidth]{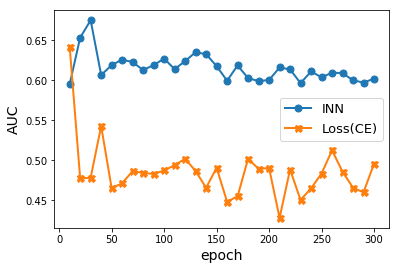}
\caption{Comparison of AUC values for clean/noisy sample classification for Clothing1M. }
\label{fig:clothing1m_auc}
\vspace{-.4cm}
\end{figure}

We consider various training epochs for $f$, $T^{\text{inn}}$, from 50 to 300, and calculate the clean/noisy classification AUC values of the training data induced by the INN for each training epoch. 
We consider two small-loss methods for the baselines, whose loss functions are the standard cross-entropy (CE) and the sum of the cross-entropy and negative entropy (CE+NE), respectively.  
We evaluate the small-loss methods' clean/noisy classification AUC values based on their per-sample losses. 
%Each cases are implemented three times and averaged AUC values are listed in Figure \ref{fig:cifar10_100_auc}.
%In particular, we show that the methods based on the \textit{memorization effect} 
%is very sensitive to the choice of a train epoch while our proposed method is stable regardless of data sets.

The results are depicted in Figure \ref{fig:cifar10_100_auc}. 
We can clearly see that the INN provides consistent and high-quality results for all considering cases regardless of training epochs. In contrast, the performances of other baselines become worse as the training epoch increases. 
That implies the INN gives a more stable and powerful performance for clean/noisy classification. 
We also repeat these experiments with another deep architecture, WideResNet28-2 \cite{zagoruyko2016wide}, and leave the results in the supplementary materials.

%\paragraph{Clothing1M} 
As for Clothing1M, we extract the weights of $h$ from the estimate of RN pre-trained by ImageNet and train only $f$ for $T^{\text{inn}}$ epochs varying from 10 to 300. 
We compare the AUC values of the INN with those based on the loss values of RN trained by CE. 
Figure \ref{fig:clothing1m_auc} shows that the INN is superior and more stable throughout the whole training procedure compared to the competitor, which again assures the effectiveness of the INN. 

%\begin{figure}[t]
%	\vskip 0.2in
%	\begin{center}
%		\centerline{
%			\includegraphics[width=0.35\textwidth]{./fig/RN50_AUC_RN50_clothing1M.png}}
%		\caption{Comparison of AUC values for clean/noisy sample classification between the training \textit{INN} scores and those induced by loss values. }
%		\label{fig:clothing1m_auc}
%	\end{center}
%	\vskip -0.2in
%\end{figure}

\subsection{INN with heavy noisy rates}
\label{sec:exp-heavy_noise}

We experiment with situations where training data are highly contaminated with noisy labels. 
We analyze CIFAR10\&100 and change their ground-truth labels with a high probability. 
Like the previous analysis, we compare the INN to the two small-loss-based methods trained with the CE and CE+NE, respectively. 
%We compare their clean/noisy classification AUC values from the training data, and vary the total training epochs of $f$, $T^{\text{inn}}$ from 10 to 300

The best clean/noisy classification AUC values of the training data for each method are summarized in Table \ref{table:inn-heavy_noise}.
In the heavy noise case, the proportion of the clean labeled data is not large. 
Thus, reducing the loss of clean labeled data may not be an optimal direction to reduce the overall loss values in the early learning stages, leading to the degradation of the small-loss strategy. 
In contrast, the INN can still identify clean data from noisy ones effectively even when many noisy labels exist and outperform the small-loss methods with large margins.
We conjecture that this performance differences arise because the consistency effect is more insensitive to the number of noisy labels. %, and we will leave the issue to future work. 

\begin{table}[t]
\fontsize{8.2pt}{8.2pt}
\selectfont
\centering
\caption[9pt]{The best clean/noisy classification AUC values of the INN and small-loss methods. 
The averaged values over three trials are listed, and standard deviations are also given in the parenthesis.
%The test accuracies of each method at the last training epoch are also reported in the parentheses. 
%The results marked with * are re-implemented by us and the other results except for INN-DivideMix are copied from the comparison table in \cite{li2020dividemix} or their original papers. 
%and the results and  Another marked ones with ** denote the results from the original papers. For the others, we copied the reported ones from the DivideMix \cite{li2020dividemix}.
}
\label{table:inn-heavy_noise}
\vskip 0.11in
\begin{center}
\begin{tabular}{l|c|c|c|c|c}
\toprule
{Data set}            & \multicolumn{2}{c|}{CIFAR10} & \multicolumn{3}{c}{CIFAR100} \\ 
\midrule
\midrule
{Noise type}        & \multicolumn{2}{c|}{Symm.}    & \multicolumn{2}{c|}{Symm.} & Asymm. \\ 
\midrule
{Noise rate ($r$)}    & 0.8          & 0.9          & 0.8       & 0.9     & 0.4         \\ 
\midrule
\midrule
CE    & 0.857 (0.011)       & 0.756 (0.015)        & 0.809 (0.005)     & 0.690 (0.009) & 0.589 (0.002) \\
\midrule
CE+NE  & 0.854 (0.012)         & 0.750 (0.017)         & 0.807 (0.015)      & 0.695 (0.011)  & 0.608 (0.003)     \\
\midrule
\midrule
INN  & {\bf 0.885} (0.014)         &  {\bf 0.817} (0.018)         & {\bf 0.853} (0.016)      & {\bf 0.717} (0.013)    & {\bf 0.671} (0.005)   \\
\bottomrule
\end{tabular}
\end{center}
\vskip -0.11in
\end{table}

\subsection{Analysis of imbalanced data}
\label{sec:exp-imb_data}

We also analyze the noisy data where the ground-truth labels are imbalanced. 
In this section, we consider the two-class classification task.
From CIFAR10, we randomly sample two classes: first and second labels are regarded as a majority class and a minority class, respectively. 
We gather all images in the first class and 10\% randomly sampled images in the second class. 
We relabel the majority and minority classes to 0 and 1, respectively, i.e. $y^*\in\{0,1\}$, and for each sample, we flip its label with a probability of 0.3 to generate training data with noisy labels.
%Then for each sample, we flip its label with a probability of 0.3 to generate training data.
We compare the normalized score distributions of the INN and the small-loss method (CE) with the training data. 
For each distribution, we make four histograms by considering two factors: 1) whether the ground-truth label is 0 or 1 and 2) whether the observed label is 0 or 1. 

\begin{figure*}[t]
\vskip 0.2in
\begin{center}
\centerline{
\includegraphics[width=0.249\textwidth]{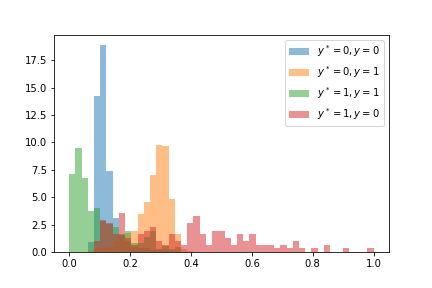}
\includegraphics[width=0.249\textwidth]{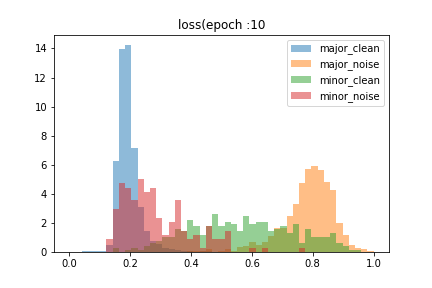}
\includegraphics[width=0.249\textwidth]{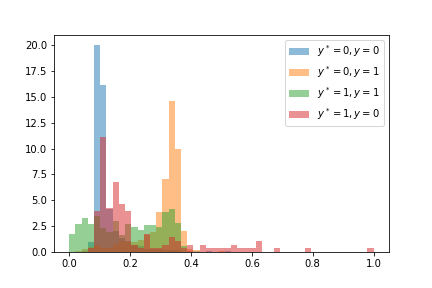}
\includegraphics[width=0.249\textwidth]{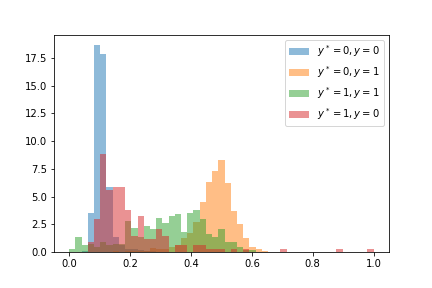}
}
\centerline{
\includegraphics[width=0.249\textwidth]{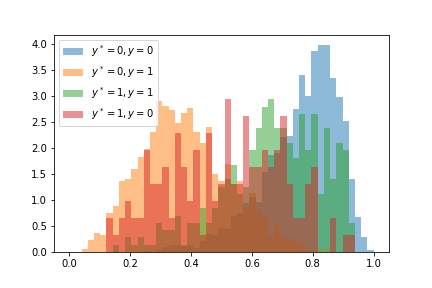}
\includegraphics[width=0.249\textwidth]{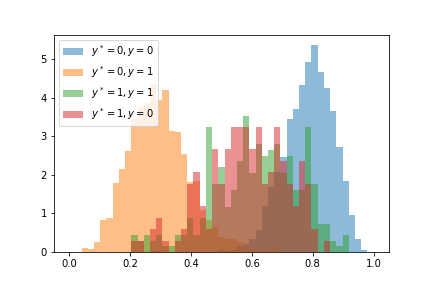}
\includegraphics[width=0.249\textwidth]{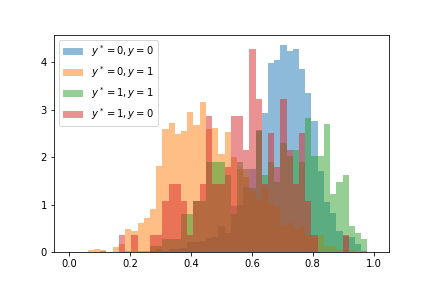}
\includegraphics[width=0.249\textwidth]{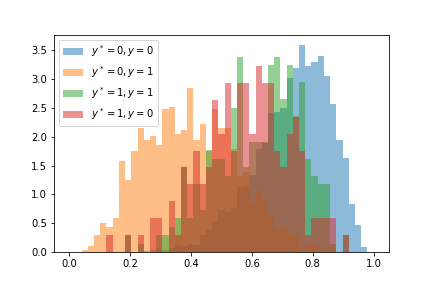}
}
\caption{Score histograms for {\bf (Upper)} the small-loss strategy (CE) and {\bf (Lower)} the INN for imbalanced data. 
We fix the number of training epochs to 50 for the INN and 10 for the small-loss method.
{(\bf Left to Right)} We pick four class pairs, $(1,2), (3,8),(4,5)$ and $(7,9)$ and the first class for each pair is treated as a majority class. 
Data with $(y^*,y)=(0,0),(0,1),(1,1)$ and $(1,0)$ are colored by blue, orange, green and red, respectively. 
%For each panel, we illustrate four histograms by considering two factors: 1) whether the ground-truth is the majority label or not and 2) whether the observed label is clean or noisy. 
}
\label{fig:imb_data}
\end{center}
\vskip -0.2in
\end{figure*}

As can be seen in Figure \ref{fig:imb_data}, we observe that the model trained by the standard CE does not always  prioritize memorizing the clean-labeled data anymore, i.e. data with $y=y^*$, rather memorizing the data with $y=0$, which implies the memorization effect may not occur in these imbalanced cases. 
Furthermore, their loss distributions are very unstable with respect to the training epochs. 
We report the histograms at other training epochs in the supplementary materials. 
In contrast, the consistency effect is still observable though not clear. 
The INN can separate clean samples from noisy ones to some extent, even if they belong to the minor class. 
%In the following section, we will also claim that this difference between ours and the small-loss method get amplified when we construct prediction models based on their results, respectively. 

\subsection{Constructing noise-robust classifiers}
\label{sec:exp-inn_dividemix}

\begin{table}[t]
\fontsize{8.2pt}{8.2pt}
\selectfont
\centering
\caption[9pt]{Comparison of the test accuracies(\%) of various method. 
%The averaged values (and standard deviations) over three trials are listed. 
The results marked with * are re-implemented by us and the other results except for INN$+$DivideMix are copied from the comparison table in \cite{li2020dividemix} or their original papers. 
We only report the best results and attach the last accuracy results in the supplementary materials. 
}
\label{table:inndividemix-heavy_noise}
\vskip 0.11in
\begin{center}
\begin{tabular}{l|c|c|c|c|c}
\toprule
{Data set}            & \multicolumn{2}{c|}{CIFAR10} & \multicolumn{3}{c}{CIFAR100} \\ 
\midrule
\midrule
{Noise type}        & \multicolumn{2}{c|}{Symm.}    & \multicolumn{2}{c|}{Symm.} & Asymm. \\ 
\midrule
{Noise rate ($r$)}    & 0.8          & 0.9          & 0.8       & 0.9     & 0.4         \\ 
\midrule
\midrule
Cross-Entropy    & 62.9        & 42.7         & 19.9      & 10.1  & 42.7  \\
%Forward \textit{T} \cite{patrini2017making}    & - (-)       & - (-)        & - (-)     & - (-) & 70.8 (-) \\
Co-teaching$+$ \cite{pmlr-v97-yu19b}   & 67.4        & 47.9         & 27.9      &  13.7 & -  \\
P-correction \cite{yi2019probabilistic}    & 77.5        & 58.9         & 31.1      &  15.3  &  - \\
MLNT \cite{li2019learning}    & -        & 59.1 (1.12)        & -      & -  & -  \\
M-correction \cite{arazo2019unsupervised}    & 86.8        & 69.1         & 48.2      & 24.3  & -  \\
\midrule
DivideMix \cite{li2020dividemix}  & 93.2          & 76.0          & {\bf 60.2}       & 31.5   & -     \\
DivideMix*  & 92.90 (1.08)         & 71.34 (1.43)         & 58.26 (1.01)      & 31.36 (0.67)  & 59.26 (1.08)     \\
\midrule
\midrule
INN$+$DivideMix  & {\bf 93.48} (1.01)         & {\bf 81.20} (1.05)         & 59.04 (0.98)      & {\bf 33.11} (0.82)    & {\bf 63.04} (1.10)   \\
\bottomrule
\end{tabular}
\end{center}
\vskip -0.11in
\end{table}

\begin{table}[t]
\fontsize{8.2pt}{8.2pt}
\selectfont
\centering
\caption[9pt]{The test accuracies(\%) of the DivideMix and the modified DivideMix with the INN.  
The averaged values and standard deviations are reported.
}
\label{table:inndividemix-imb}
\vskip 0.11in
\begin{center}
\begin{tabular}{l|cc|cc}
\toprule
Data set& \multicolumn{4}{c}{Imbalanced CIFAR10} \\
\midrule
Classes        & \multicolumn{2}{c|}{DivideMix} & \multicolumn{2}{c}{INN$+$DivideMix} \\ 
&Best&Last&Best&Last\\
\midrule
% 0 and 6   & 91.91 (-)          & 83.82 (-)   & 92.87 (-)          & 85.05 (-)   \\
 1 and 2   & 89.23 (0.40)          & 82.21 (0.57)   & {\bf 92.82} (0.27)       & {\bf 87.75} (0.77)   \\
 3 and 8   & 87.49 (0.68)       & 82.24 (1.65)   & {\bf 9
 3.02} (0.12)          & {\bf 91.06} (0.18)  \\ 
 4 and 5   & 79.41 (0.68)          & 78.35 (0.98)   & {\bf 82.82} (0.20)          & {\bf 82.79} (0.13)   \\ 
 7 and 9   & 85.64 (0.35)          & 79.80 (1.92)   & {\bf 86.33} (0.45)          & {\bf 84.15} (0.25)   \\ 
\bottomrule
\end{tabular}
\end{center}
\vskip -0.11in
\end{table}

%In Section \ref{sec:app_inn}, we have mentioned that our method is also helpful for learning deep classification models with high performance with noisy training data.
The INN is also helpful for learning deep classification models with high performance with noisy training data. 
%We give a simple but powerful application of the INN to construct deep prediction models robust to noisy training data.
Many conventional learning frameworks built on the small-loss strategy usually begin with learning models with the whole training data by minimizing the standard loss function, such as CE, for a few epochs. 
Due to the memorization effect, the early estimated models tend to memorize the clean-labeled training data first. 
Initialized with the estimated models, they conduct their own strategies to train models. 
After each training epoch, they update the per-sample loss values with the current model and utilize this information at the next training epoch.

We can modify them by simple modifications with the INN. 
First, we replace their initialized models with new ones trained with the INN.
%With the INN scores, we separate the training data to labeled and unlabeled data by fitting a two-component mixture model to the INN scores of the training data. 
We fit a two-component mixture model to the INN scores of the training data and split the data into the labeled and unlabeled data based on the posterior probability of belonging to the clean cluster (the cluster with a larger mean). 
%We regard samples whose posterior probabilities of belonging to the clean cluster (the cluster with a larger mean) is larger than 0.5 as labeled data and treat the remained samples as unlabeled data by discarding their labels. 
With the labeled and unlabeled data, we train prediction models for a few epochs using a semi-supervised learning method, such as the MixMatch \cite{NEURIPS2019_1cd138d0}, then use them as the initialized estimates. 
Second, after each training epoch, we recalculate the INN scores with the current prediction model and utilize these score values instead of the loss values at the next training epoch. 
%Second, we use the INN score's information instead of that induced by the loss values at each training epoch. 

In this work, we provide an example to mix the INN and the DivideMix \cite{li2020dividemix}, known as one of the state-of-the-art methods to learning models with data mixed with noisy labels. 
We add a detailed algorithm of this combination in the supplementary materials. 
%In this work, we give an example to mix the INN and the DivideMix method, known as one of the state-of-the-art methods \cite{li2020dividemix} to learning models with data mixed with noisy labels. 
%For detailed algorithm of the combination of the DivideMix and the INN, see the supplementary materials. 
We again stress that any other small-loss-based learning frameworks than the DivideMix are also available to be combined with the INN to train better prediction models, and that there are many rooms to develop our simple application. 
%We refer \cite{li2020dividemix} for the hyperparameter settings, e.g. the tuning parameter in the loss function, of the SSL phase of the DivideMix. 

To assess the prediction performance of our modification, we analyze the same noisy data sets in Section \ref{sec:exp-heavy_noise} and \ref{sec:exp-imb_data}. 
We carry out test accuracy comparison with the modified DivideMix with the INN (INN$+$DivideMix) and other baselines, which are reported in Table  \ref{table:inndividemix-heavy_noise} and \ref{table:inndividemix-imb}.\footnote{We complete our source code based on the public GitHub code of the DivideMix.} 
%(For other cases, see the supplementary materials). 
We can check that our modified learning framework works better than the existing methods, including the DivideMix, in the cases where the training labels are highly polluted or imbalanced. And the performance gap between the best and last models trained with our method is smaller than the original DivideMix, which means the INN makes learning procedures more stable. 
We also make additional accuracy tests for data sets with noise rates not too severe and observe that ours gives no significant improvements since the DivideMix already works well. 
We report these results in the supplementary materials. 
%First, we can check that our modified learning framework works better than the existing methods including the DivideMix in heavy noise cases. 
%That means the INN provides more desirable initialization than the small-loss strategy. 
%In addition, when analyzing noisy data with imbalanced label distribution, the INN is superior to the small-loss competitor with large margins, and this is because the INN can identify the clean data belonging to the minor class better. 

\subsection{Ablation study}
\label{sec:ablation}

We empirically investigate how the choices of dissimilarity measure $h$ and the loss function $l$ affect the INN, whose detailed analyses are described in the supplementary materials. 
We observe that the Euclidean distance on the penultimate layer of a DNN trained by the training data is the best dissimilarity measure. And for the loss function, the MixUp objective function yields the best results. 
%We attach the detailed analyses in the supplementary materials. 
%We need the results about nearest neighborhoods for each sample, our method takes more time than the competitor, which could be a limitation of ours. 

We also evaluate the total training time of the INN and compare it to that of the standard small-loss method. % on CIFAR10.
We use a single NVIDIA TITAN XP GPU, and the results are in the supplementary materials.
It takes much more time to implement the INN than the competitor since it needs to extract the nearest neighbors for each sample, which could be one of our limitations. 
More study to lighten the computation burden of the INN is required. 

\section{Concluding remarks}
\label{sec:conclusions}

In this study, we proposed a new and novel approach, called the INN, to identify clean labeled samples from training data with noisy labels based on a new finding called the consistency effect that discrepancies of predictions at neighbor regions of clean and noisy data are consistently observed.
We empirically demonstrated that the INN is stable and superior even when the training labels are heavily contaminated or imbalanced. 
%and that simply unifying the INN and existing small-loss-based learning methods can construct better prediction models.  

It would be interesting to apply our methods to unsupervised anomaly detection problems \cite{7410534,rda,dagmm,goad}. 
After annotating labels (normal or abnormal) to the training data in a certain way, we can regard the task as the two-class noisy label problem. 
We expect that our methods would solve the anomaly detection problems successfully. 

%\textbf{Broader impact} 
%It is a substantial issue in AI fields to collect a large amount of clean labeled data to train deep learning models, and the data labeling for AI is becoming a huge market. 
%Much of the labeling work is done by human experts in place with cheap labor, but it is still difficult, expensive, and time-consuming to gather clean labels. 
%With the INN, we can refine clean labeled data from noisy labeled data effectively and inexpensively. 
%Thus, our study has a potentially positive impact on the growth of the data labeling market. 
%At the same time, this work may negatively affect society because cost-cutting for data labeling procedures results in a decline in employment. 

\bibliography{inn_reference}
\bibliographystyle{unsrt}

\newpage

\section{Supplementary materials for the INN}

\title{\bf Supplementary materials for\\ \enquote{INN: A Method Identifying Clean-annotated Samples via Consistency Effect in Deep Neural Networks}}
\maketitle

\appendix

\section{Proof of Lemma 1}
\label{sec:proof}
Let $\bz=(\bx,y)$ and $\tilde{\bX}=\{\tilde{\bx}_1,\ldots,\tilde{\bx}_L\}$ be a given training sample and a set of its nearest training inputs, respectively. 
Since the model $f$ perfectly over-fits the MixUp loss function, it is linear over between any two given training inputs. So, for $l=1,\ldots,L$ the following equality holds:
\bean
\int_0^1 f_y(\alpha\bx+(1-\alpha\tilde{\bx}_l);\hat{\theta})d\alpha=\left\{
\begin{array}{ll}
1& \text{if }y=\tilde{y}_l,\\
\frac{1}{2}& \text{otherwise,}
\end{array}
\right.
\eean
where $\tilde{y}_l$ is the corresponding observed label of $\tilde{\bx}_l$. Hence, the INN score can be rewritten as:
\bea
\label{formula1}
I(\bz)=\frac{1}{L}\sum_{l=1}^L \int_0^1 f_y(\alpha\bx+(1-\alpha\tilde{\bx}_l);\hat{\theta})d\alpha=\frac{1}{L}\sum_{l=1}^L \left(  I(y=\tilde{y}_l)+\frac{1}{2}I(y\neq\tilde{y}_l)\right).
\eea
By using $\text{argmax}_{k\in[K]} \sum_{l=1}^L I(\tilde{y}_l=k)=y^*$, we also have 
\bea
\label{formula2}
\sum_{l=1}^L I(y^*=\tilde{y}_l)\ge \frac{L}{K} \text{ and } 
\sum_{l=1}^L I(y^*\neq\tilde{y}_l)\le \frac{L-K}{K}.
\eea 
With (\ref{formula1}) and (\ref{formula2}), 
\bea
\label{ineq1}
I(\bz) \ge \frac{1}{L} \left[ \frac{L}{K}+\frac{L-K}{2K} \right]
\ge \frac{1}{L}\left( \frac{3L}{2K}-\frac{1}{2} \right)
\eea
if $\bz$ is cleanly labeled, i.e. $y=y^*$, and 
\bea
\label{ineq2}
I(\bz) &=& \frac{1}{L}\sum_{l=1}^L \left(  I(y=\tilde{y}_l)+\frac{1}{2}I(y\neq\tilde{y}_l)\right)\nonumber\\
&\le& \frac{1}{L}\sum_{l=1}^L \left( I(y^*\neq \tilde{y}_l) +\frac{1}{2} \right)
\le \frac{1}{L}\left(\frac{L}{K}-\frac{1}{2}\right)
\eea
if $\bz$ is noisy labeled. 
From (\ref{ineq1}) and (\ref{ineq2}), we have the following inequality and the proof is completed:
\bean
\underset{\bz\in\cC^{tr}}{\min}I(\bz)-\underset{\bz\in\cN^{tr}}{\max}I(\bz)\ge \frac{1}{2K}>0. \text{\qed}
\eean
%thus the proof is done.

\section{INN with another architecture}

We carry out additional experiments for the performance test of the INN using another architecture. 
Here, we consider WideResNet28-2 \citep{zagoruyko2016wide} and we follow the main manuscript's implementation settings, such as the strategies to impose noisy labels and learning schedules. 

\begin{figure}[t]
\renewcommand\thefigure{B.1} 
\vskip 0.2in
\begin{center}
\centerline{
\includegraphics[width=0.245\textwidth]{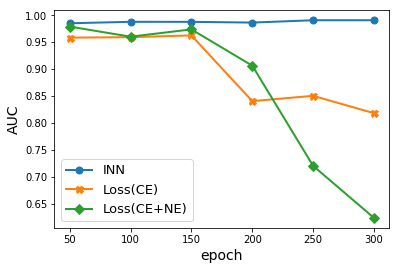}
\includegraphics[width=0.245\textwidth]{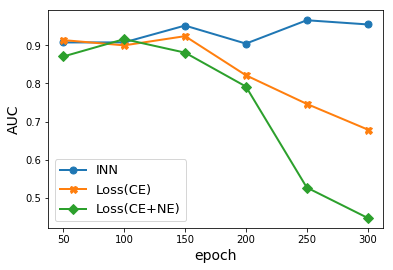}
\includegraphics[width=0.245\textwidth]{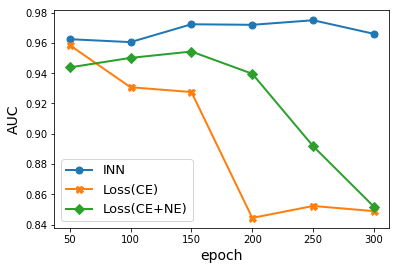}
\includegraphics[width=0.245\textwidth]{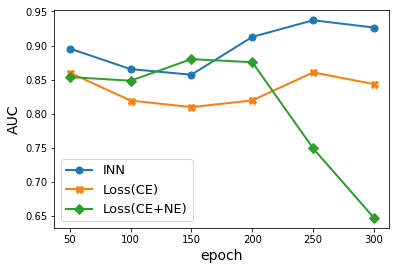}
}
\centerline{
\includegraphics[width=0.245\textwidth]{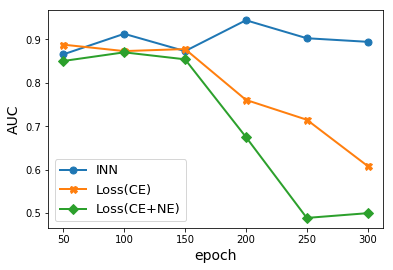}
\includegraphics[width=0.245\textwidth]{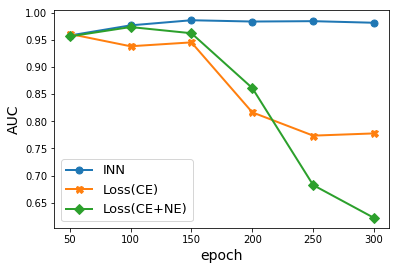}
\includegraphics[width=0.245\textwidth]{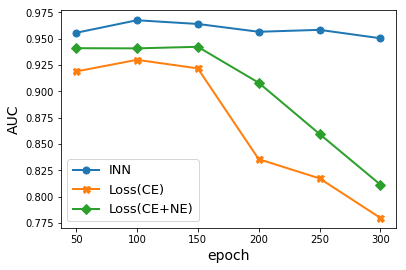}
\includegraphics[width=0.245\textwidth]{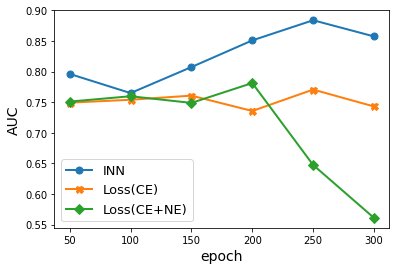}
}

\caption{Comparison of AUC values for clean/noisy sample classification between the INN and the two small-loss methods. We utilize the WideResNet28-2 architecture. The plot positioned at the $i$-th row from the top and the $j$-th column from the left is the result of the Case$i$-$j$. As for the description of the eight cases, see the main manuscript.
}
\label{fig:cifar10_100_auc_wrn_sgd}
\end{center}
\vskip -0.2in
\end{figure}

We depict the results in Figure \ref{fig:cifar10_100_auc_wrn_sgd}. 
Similar to the results with the PRN architecture, the clean/noisy classification performances of the INN are insensitive to the choice of the training epochs and consistently outperform other two competitors based on loss values.

\section{Instability of the small-loss method for imbalanced data}
%\section{Imbalanced data analysis with INN}

\begin{figure}[t]
\renewcommand\thefigure{C.1} 
\vskip 0.2in
\begin{center}
\centerline{
\includegraphics[width=0.245\textwidth]{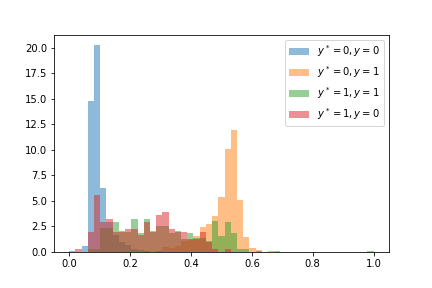}
\includegraphics[width=0.245\textwidth]{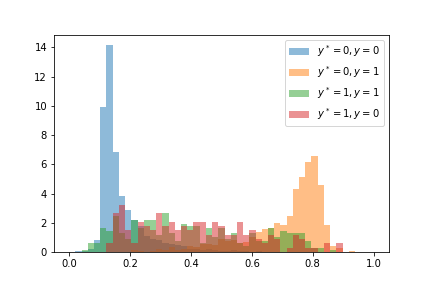}
\includegraphics[width=0.245\textwidth]{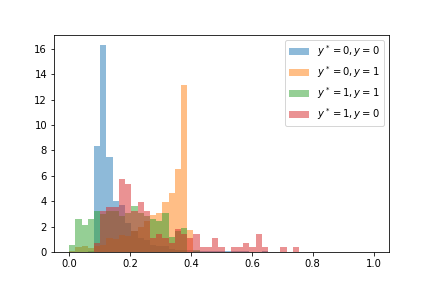}
\includegraphics[width=0.245\textwidth]{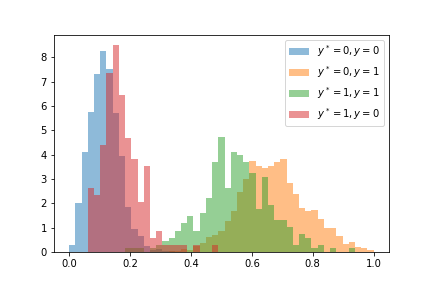}
}
%\centerline{
%\includegraphics[width=0.245\textwidth]{./supp_fig/small_loss_[1, 2]_ep(3)_density_True.png}
%\includegraphics[width=0.245\textwidth]{./supp_fig/small_loss_[3, 8]_ep(3)_density_True.png}
%\includegraphics[width=0.245\textwidth]{./supp_fig/small_loss_[4, 5]_ep(3)_density_True.png}
%\includegraphics[width=0.245\textwidth]{./supp_fig/small_loss_[7, 9]_ep(3)_density_True.png}
%}

\centerline{
\includegraphics[width=0.245\textwidth]{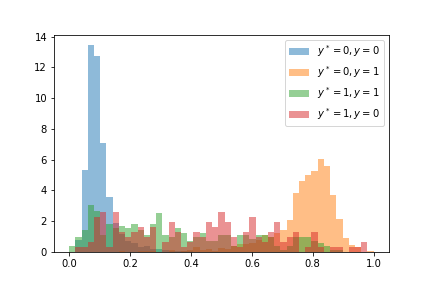}
\includegraphics[width=0.245\textwidth]{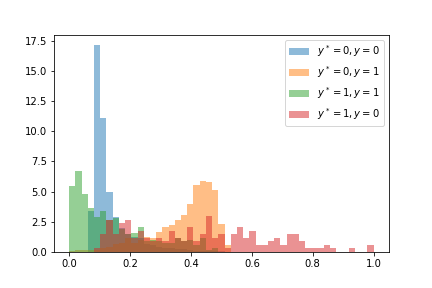}
\includegraphics[width=0.245\textwidth]{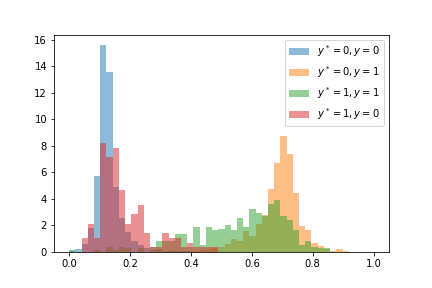}
\includegraphics[width=0.245\textwidth]{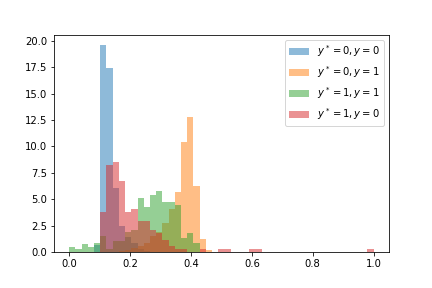}
}

\centerline{
\includegraphics[width=0.245\textwidth]{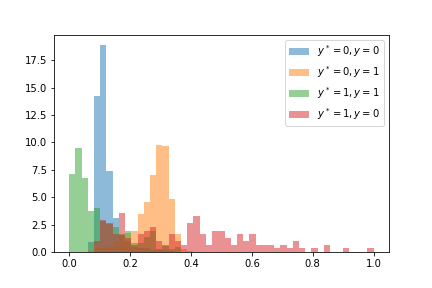}
\includegraphics[width=0.245\textwidth]{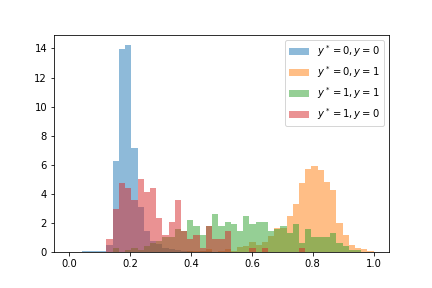}
\includegraphics[width=0.245\textwidth]{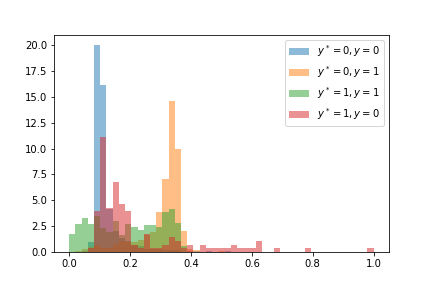}
\includegraphics[width=0.245\textwidth]{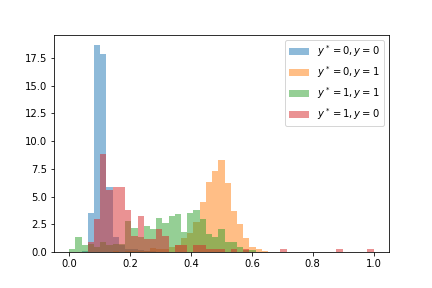}
}

\centerline{
\includegraphics[width=0.245\textwidth]{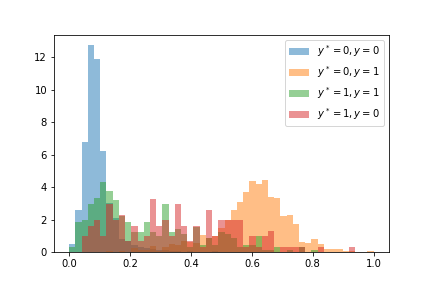}
\includegraphics[width=0.245\textwidth]{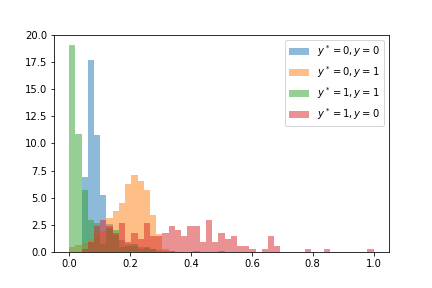}
\includegraphics[width=0.245\textwidth]{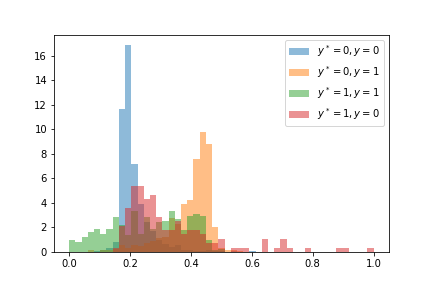}
\includegraphics[width=0.245\textwidth]{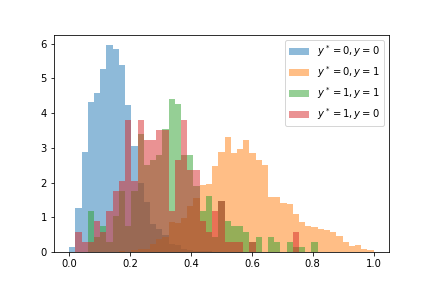}
}
\centerline{
\includegraphics[width=0.245\textwidth]{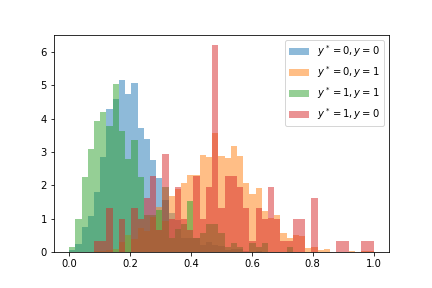}
\includegraphics[width=0.245\textwidth]{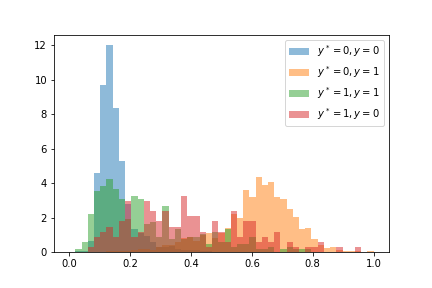}
\includegraphics[width=0.245\textwidth]{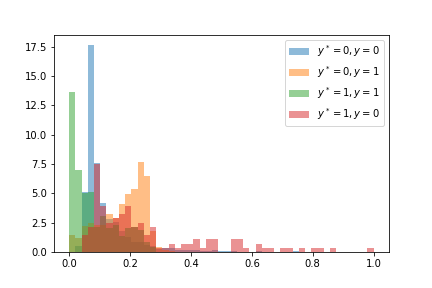}
\includegraphics[width=0.245\textwidth]{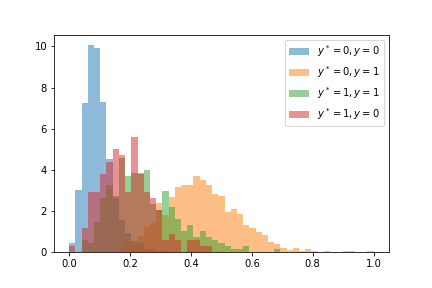}
}

\caption{The histograms of the loss value. ({\bf Upper to Lower}) We consider five training epochs (1,5,10,15 and 20). ({\bf Left to Right}) We pick four class pairs, $(1,2), (3,8), (4,5)$ and $(7,9)$, and the first class for each pair is treated as a majority class. Data with $(y^*,y)=(0,0),(0,1),(1,1)$ and $(1,0)$ are colored by blue, orange, green and red, respectively.
}
\label{fig:imbal_loss_hist}
\end{center}
\vskip -0.1in
\end{figure}

Figure \ref{fig:imbal_loss_hist} illustrates histograms of the loss values at various training epochs when we analyze imbalanced data. 
The same data sets used in Section 4.4 of the main manuscript are considered. 
We can check that the loss distribution is unstable, so it is hard to choose an optimal training epoch. 

\section{Modification of DivideMix with INN}
\subsection{Detailed algorithm description}

Here, %we explain the way how to modify the DivideMix method with the INN. 
we describe our modification (INN$+$DivideMix) in detail by comparing it with the original DivideMix algorithm.
%Before we describe the way we combine the INN with the DivideMix method, we give a brief review of the DivideMix. 
%The DivideMix consists of two steps. 
%First, two prediction models are trained with the whole training data for a few epochs by minimizing the sum of the cross-entropy and negative entropy. 

%Second, starting from the two current estimated models, the DivideMix fits a two-component Gaussian mixture model (GMM) on a per-sample loss distribution, then split the training data into labeled and unlabeled data based on the posterior probability belonging to the clean cluster (the cluster with a smaller mean). 
%To be more specific, we treat the data whose posterior probability of the clean cluster is larger than 0.5 as the labeled data and the remaining data as the unlabeled data by discarding their corresponding labels.
%With the labeled and unlabeled data, 

\subsubsection{Step 1}
\paragraph{Original DivideMix}
The DivideMix trains two prediction models by minimizing the sum of the cross-entropy and negative entropy with two initial independent parameters for a pre-specified training epoch using the whole training data. 
The training epoch usually ranges from 10 to 30, depending on a data set. 
This part aims to get two initialization models robust to noisy labels to some extent via the memorization effect. 
%The two estimated models are then used as initialization of two prediction models in the following SSL phase. 

\paragraph{INN$+$DivideMix}
First, we conduct the INN method twice to obtain two corresponding INN score sets of the training data with random initializations. 
With each INN score set, we separate the training data to labeled and unlabeled data by fitting a two-component Beta Mixture Model (BMM) to the INN scores using the Expectation-Maximization algorithm. 
We regard samples whose posterior probabilities of belonging to the clean cluster (the cluster with a larger mean) is larger than 0.5 as labeled data and treat the remained samples as unlabeled data by discarding their labels. 
Then, we utilize a semi-supervised learning (SSL) method to train two prediction models for a pre-specified training epoch with each pair of the labeled and unlabeled data. 
Any SSL method can be applied, and in our experiments, we adopt the MixMatch \citep{berthelot2019mixmatch}. 
%, which will be used as initialized models for the following SSL phase. 

\subsubsection{Step 2}
\paragraph{Original DivideMix}
The DivideMix updates two prediction models pre-trained from the first step. 
For each prediction model, the DivideMix fits a two-component Gaussian Mixture Model (GMM) on a per-sample loss distribution of the model to divide the training set into a labeled data set and an unlabeled data set, which is called the \textit{co-divide} procedure. 
Then, the two models exchange the \textit{co-divided} data sets and are trained based on the exchanged data sets by use of the SSL method modifying the {MixMatch} with the \textit{co-refinement} and the \textit{co-guessing} techniques. 
The DivideMix iterates the \textit{co-divide} and the modified MixMatch method alternately for a pre-specified training epoch to construct the final prediction models. 

\paragraph{INN$+$DivideMix}
Similar to the original DivideMix, our method starts with two initialized prediction models from the first step of ours. 
For each prediction model, we calculate the INN score set. 
Note that we need a prediction model $f$ and a feature model $h$ to conduct the INN. 
Here, we set $f$ and $h$ to the current prediction model and its penultimate layer, i.e., the highest hidden layer, respectively. 
With each INN score set, we fit a two-component Beta Mixture Model (BMM) and split the training data into labeled and unlabeled data by using the posterior probabilities. 
Then, we utilize the same SSL algorithm used in the original DivideMix with the two pairs of labeled and unlabeled data sets to update the two prediction models. 
We repeat the above learning procedure for a pre-specified training epoch. 
We summarize our modification in Algorithm \ref{alg:inn_dividemix}. 
Algorithm \ref{alg:inn_dividemix} requires three kinds of training epochs, $T_1,T_2,$ and $T_3$.  
In practice, we set $(T_1,T_2,T_3)=(30,30,9)$ for all cases. 

\renewcommand{\thealgorithm}{D.1}
\begin{algorithm}[tb]
	\caption{INN+DivideMix: line 1-9: Step 1, line 10-20: Step 2}
	\label{alg:inn_dividemix}
	\begin{algorithmic}[1]
		\INPUT Training data $\cD^{\text{tr}}=\{(\bx_1,y_1),\ldots,(\bx_n,y_n)\}$, two prediction models $g_1$ and $g_2$, three integers $T_1,T_2$ and $T_3$, an optimizer $\cO$
		%\FOR {$t=1$ {\bfseries to} $T^{\text{wu}}$}
		%\STATE $\hat{\phi}^{(1)},\hat{\phi}^{(2)}\leftarrow Sup(g(\cdot;\phi^{(1)},g(\cdot;\phi^{(2)}),\cD^{tr},\cO)$ \hfill \algorithmiccomment //Supervised learning
		%\ENDFOR
		\STATE Train two prediction models $f_1$ and $f_2$ and two feature models $h_1$ and $h_2$.
		\STATE $\cS_1\leftarrow\{s_{11},\ldots,s_{1n}\}$; $\cS_2\leftarrow\{s_{21},\ldots,s_{2n}\}$ \hfill  //two INN score sets with $(f_1,h_1)$ and $(f_2,h_2)$, respectively
		\STATE $\cW_1=\{w_{11},\ldots,w_{1n}\}\leftarrow \text{BMM}(\cS_1)$; $\cW_2=\{w_{21},\ldots,w_{2n}\}\leftarrow \text{BMM}(\cS_2)$ \hfill  //clean posterior probabilities
		\STATE $\cL_1^{\text{tr}}=\{(\bx_i,y_i):w_{1i}\ge 0.5,i=1,\ldots,n\}$; 
		$\cU_1^{\text{tr}}=\{\bx_i:w_{1i}< 0.5,i=1,\ldots,n\}$ 
		\STATE $\cL_2^{\text{tr}}=\{(\bx_{i},y_i):w_{2i}\ge 0.5,i=1,\ldots,n\}$; 
		$\cU_2^{\text{tr}}=\{\bx_i:w_{2i}< 0.5,i=1,\ldots,n\}$ \\\hfill  //two pairs of labeled and unlabeled data
		\FOR {$t_1=1$ {\bfseries to} $T_1$}
		\STATE $g_1\leftarrow\text{MixMatch}(g_1,\cL_2^{\text{tr}},\cU_2^{\text{tr}},\cO)$
		\STATE $g_2\leftarrow\text{MixMatch}(g_2,\cL_1^{\text{tr}},\cU_1^{\text{tr}},\cO)$ \hfill  //conduct the MixMatch
        \ENDFOR
		\FOR {$t_2=1$ {\bfseries to} $T_2$}
		\STATE $f_1\leftarrow g_1$; $h_1\leftarrow \text{penult}(g_1)$
		\hfill  //update $f_1$ and $g_1$
		\STATE $f_2\leftarrow g_2$; $h_2\leftarrow \text{penult}(g_2)$ \hfill  //update $f_2$ and $h_2$
		\STATE $\cS_1\leftarrow\{s_{11},\ldots,s_{1n}\}$; $\cS_2\leftarrow\{s_{21},\ldots,s_{2n}\}$ \hfill  //two INN score sets 
		\STATE $\cW_1=\{w_{11},\ldots,w_{1n}\}\leftarrow \text{BMM}(\cS_1)$; $\cW_2=\{w_{21},\ldots,w_{2n}\}\leftarrow \text{BMM}(\cS_2)$ \hfill  //clean posterior probabilities
		\STATE $\cL_1^{\text{tr}}=\{(\bx_i,y_i):w_{1i}\ge 0.5,i=1,\ldots,n\}$; 
		$\cU_1^{\text{tr}}=\{\bx_i:w_{1i}< 0.5,i=1,\ldots,n\}$ 
		\STATE $\cL_2^{\text{tr}}=\{(\bx_{i},y_i):w_{2i}\ge 0.5,i=1,\ldots,n\}$; 
		$\cU_2^{\text{tr}}=\{\bx_i:w_{2i}< 0.5,i=1,\ldots,n\}$ \\\hfill  //two pairs of labeled and unlabeled data
		\FOR {$t_3=1$ {\bfseries to} $T_3$}
        \STATE $g_1,g_2\leftarrow\text{DivideMix-SSL}(g_1,g_2,\cL_1^{\text{tr}},\cL_2^{\text{tr}},\cU_1^{\text{tr}},\cU_2^{\text{tr}},\cO)$ \hfill  //conduct the SSL part of the DivideMix
		\ENDFOR
		\ENDFOR
		\OUTPUT $g_1$ and $g_2$ \hfill  //final outputs
	\end{algorithmic}
\end{algorithm}

\subsection{Performance tests of INN$+$DivideMix}

We report the last test accuracy results of the INN$+$DivideMix on severely contaminated data sets. 
Table \ref{table:inndividemix-heavy_noise_last} shows that the modified DivideMix with the INN improves the DivideMix with large margins.

The prediction performance of the INN in ordinary cases where there are not many noisy labeled samples in the training data are provided in Table \ref{table:inndividemix-ordinal_best}. 
Our modification does not give visible enhancement in these situations since the DivideMix already works well, so there is not much room to develop the original one.

\begin{table}[t]
\fontsize{8.2pt}{8.2pt}
\selectfont
\renewcommand\thetable{D.1} 
\centering
\caption[9pt]{Comparison of the last test accuracies(\%) of various methods on severely polluted data sets. 
The averaged values (and standard deviations) are listed. 
The results marked with * are re-implemented by us and the other results except for INN$+$DivideMix are copied from the comparison table in \cite{li2020dividemix} or their original papers. 
}
\label{table:inndividemix-heavy_noise_last}
\vskip 0.11in
\begin{center}
\begin{tabular}{l|c|c|c|c|c}
\hline
{Data set}            & \multicolumn{2}{c|}{CIFAR10} & \multicolumn{3}{c}{CIFAR100} \\ 
\hline
\hline
{Noise type}        & \multicolumn{2}{c|}{Symm.}    & \multicolumn{2}{c|}{Symm.} & Asymm. \\ 
\hline
{Noise rate ($r$)}    & 0.8          & 0.9          & 0.8       & 0.9     & 0.4         \\ 
\hline
\hline
Cross-Entropy    &  26.1         & 16.8         & 8.8       & 3.5  & -  \\
%Forward \textit{T} \cite{patrini2017making}    & - (-)       & - (-)        & - (-)     & - (-) & 70.8 (-) \\
Co-teaching$+$ \citep{pmlr-v97-yu19b}   & 45.5          & 30.1         & 15.5      &  8.8 & -  \\
P-correction \citep{yi2019probabilistic}    & 76.5         & 58.2         & 20.7       &  8.8  &  - \\
MLNT \citep{li2019learning}    & -        &-        & -      & -  & -  \\
M-correction \citep{arazo2019unsupervised}    & 86.6         & 68.7         & 47.6       & 20.5  & -  \\
\hline
DivideMix \citep{li2020dividemix}  & 92.9          & 75.4          & {\bf 59.6}        & 31.0   & -     \\
DivideMix*  & 92.71(1.02)         & 68.79(1.31)         & 57.78(1.07)        & 31.16(1.21)  & 53.79(1.09)     \\
\hline
\hline
INN$+$DivideMix  & {\bf 93.04}(1.19)         & {\bf 80.51}(1.05)         & 58.81(1.22)       & {\bf 32.36}(0.86)    & {\bf 59.34}(0.94)  \\
\hline
\end{tabular}
\end{center}
\vskip -0.11in
\end{table}

\begin{table}[t]
\fontsize{7.2pt}{7.2pt}
\selectfont
\renewcommand\thetable{D.2} 
\centering
\caption[9pt]{Comparison of the best (and last) test accuracies(\%) of various methods on data sets polluted not much. 
%The averaged values (and standard deviations) over three trials are listed. 
The results marked with * are re-implemented by us and the other results except for INN$+$DivideMix are copied from the comparison table in \cite{li2020dividemix} or their original papers. 
}
\label{table:inndividemix-ordinal_best}
\vskip 0.11in
\begin{center}
%\adjustbox{max width=\textwidth}{
\resizebox{\textwidth}{!}{
\begin{tabular}{l|c|c|c|c|c|c|c|c}
\hline
{Data set}            & \multicolumn{4}{c|}{CIFAR10} & \multicolumn{4}{c}{CIFAR100} \\ 
\hline
\hline
{Noise type}        & \multicolumn{4}{c|}{Symm.}    & \multicolumn{4}{c}{Symm.}  \\ 
\hline
{Noise rate ($r$)}    & 0.2     & 0.4   & 0.5   & 0.6     & 0.2     & 0.4   & 0.5   & 0.6         \\ 
\hline
\hline
Cross-Entropy    &  86.8(82.7)   &  -    & 79.4(57.9)  &  - & 62.0(61.8) &  - & 46.7(37.3) &  - \\
%Forward \textit{T} \cite{patrini2017making}    & - (-)       & - (-)        & - (-)     & - (-) & 70.8 (-) \\
Co-teaching$+$ \citep{pmlr-v97-yu19b}   & 89.5(88.2) &  -  & 85.7(84.1) &  -   &   65.6(64.1)  & - &  51.8(45.3)      &  -  \\
P-correction \citep{yi2019probabilistic}    & 92.4(92.0) &  -  & 89.1(88.7)  &  -   & 69.4(68.1) &  -  &       57.5(56.4) &  - \\
MentorNet \citep{jiang2018mentornet} &  92.0(-)   & 89.0(-)    &    - &  -  & 73.0(-)  &  68.0(-)  &  -  &  - \\  
D2L \citep{ma2018dimensionality} &  85.1(-)   & 83.4(-)    &    - &  72.8(-)  & 62.2(-)  &  52.0(-)  &  -  &  42.3(-) \\  
MLNT \citep{li2019learning}    & 92.9(92.0) &  -  & 89.3(88.8)   &  -  & 68.5(67.7) &  -  &  59.2(58.0) & -  \\
Reweight \citep{ren2018learning} &  86.9(-)  &     - &  -   & -   & 61.3(-)   &  -  &  - &  -\\  
Abstention \citep{thulasidasan2019combating} &  93.4(-)  & 90.9 (-)   &  -  & 87.6(-)    & 75.8(-)   &  68.2(-)   &  - & 59.4(-) \\  
M-correction \citep{arazo2019unsupervised}    & 94.0(93.8) &  -  & 92.0(91.9) &  -  & 73.9(73.4) &  -  & 66.1(65.4)  &  -  \\
\hline
DivideMix \citep{li2020dividemix}  & 96.1(95.7) &  94.9(-)  & 94.6(94.4)  & 94.3(-) & {\bf 77.3}({\bf 76.9})  & {\bf 75.2}(-) & 74.6(74.2)  &  72.0(-)     \\
DivideMix*  &    96.08(95.68)   & 94.95(94.51)    &  94.81(94.13) &  94.29(93.95)  & 77.23(76.84)  & 74.92(74.12) &  74.08(73.56) & 72.58(71.99)    \\
\hline
\hline
INN$+$DivideMix  & {\bf 96.11}({\bf 95.82})    & {\bf 95.01}({\bf 94.86})  & {\bf 94.99}({\bf 94.83})  & {\bf 94.87}({\bf 94.56})    & 77.27(76.86)  & 74.67({\bf 74.14})  & {\bf 75.18}({\bf 74.83})  & {\bf 72.69}({\bf 72.21})  \\
\hline
\end{tabular}
}
%}
\end{center}
\vskip -0.11in
\end{table}

\section{Detailed descriptions of ablation studies}

\subsection{Dissimilarity measure}
\label{sec:diss_msr}

As a dissimilarity measure, we utilize the Euclidean ($L_2$) distance between the feature representations (the penultimate layer representations) generated by a pre-trained prediction model,  
%\bean
%\label{eq:diss_msr}
$D(\cdot,\cdot)=L_2(h(\cdot;\hat{\eta}),h(\cdot;\hat{\eta}))$,
%\eean
%where $\hat{\theta}$ is the parameter of the pre-trained model. 
where $h(\cdot;\hat{\eta})$ is the highest feature output of a pre-trained prediction model parametrized by $\hat{\eta}$. 
Here, $\hat{\eta}$ is trained by minimizing the standard cross-entropy (CE) function on $\cD^{\text{tr}}$. % for 150 epochs.  
As an alternative feature representation, we consider using an external training data set such as ImageNet. 
We obtain another feature representation by training $\eta$ based on the ImageNet data set ($\cD^{\text{IN}}$) and compare two representation output functions trained on 1) $\cD^{\text{tr}}$ and 2) $\cD^{\text{IN}}$, respectively by evaluating AUC values of their corresponding INN scores. 
%To estimate $\hat{\eta}$, we consider two data sets: 1) the given training data, $\cD^\text{tr}$ and 2) an external training data set such as ImageNet.
%We minimize the cross-entropy for a given data set to achieve a pre-trained model denoted by $f(\bx;\hat{\theta})$, and utilize the feature vector of the penultimate layer $h(\bx;\hat{\theta})$. 
%We then obtain the nearest neighbors for each train sample by exploiting the $L_2$ distances between $f(\bx_i;\hat{\theta})$s.
%Meanwhile, we train another model $f(\bx;\hat{\theta}^{\text{inn}})$ by minimizing the cross-entropy on the asymmetrically noisy CIFAR10 data with $r=0.4$ (anCIFAR10-4). 
%For given the prediction model at the train epoch $t$, $f(\bx;\hat{\theta}_t^{\text{inn}})$, and the set of nearest neighborhoods, we calculate the \textit{INN scores} repeatedly by each 50 train epoch, that is, $t=50,100,\ldots$. 
The integrand prediction model $f(\cdot;\hat{\theta})$ used in the formula (1) of the main manuscript is trained by minimizing CE based on $\cD^{\text{tr}}$ with the maximum training epochs 300. 
By each 50 training epoch, we calculate the INN scores on $\cD^{\text{tr}}$ with $L=10$ and evaluate the AUC values for the clean/noisy sample classification problem. 
%For estimating $\hat{\theta}$, we consider two data sets: 1) the existing data and 2) the external data. 
%Thus, in this case the existing data set would be the anCIFAR10-4 data set itself, and we choose the external one as the ImageNet data set. 
%With two given feature representations, we use INN to obtain 
%We utilize the asymmetrically noisy CIFAR10 data set with the noise rate $0.4$ (anCIFAR10-4) and consider two candidates:  $L_2$ distance over the feature space pre-trained by using 1) the existing data (anCIFAR10-4) and 2) the external data (ImageNet). 
%\item With the noisy CIFAR10 data, we train a predictive model by minimizing the standard cross-entropy. For each 50 train epoch, we calculate the integration measures by each candidates with $L=10$ and compare their corresponding the Area Under a Curve (AUC) values. 
The results are depicted in the left panel of Figure \ref{fig:inn_analysis}, which show that 
%the representation trained with $\cD^{\text{tr}}$ are more helpful than the competitor for all cases, which implies 
using feature representations trained with the given training data set ($\cD^{\text{tr}}$) gives similar results. 
%it is better to search the neighborhoods using feature representations trained on the existing data. 

%We carry out an additional experiment to investigate how the number of training epochs for $h$, $T$, affects the performance of the INN. 
%We train the two feature output functions $h(\cdot;\hat{\eta}_{100})$ and $h(\cdot;\hat{\eta}_{200})$ by minimizing CE on $\cD^{\text{tr}}$ for $T=100$ and $T=200$ training epochs, respectively, and compare their corresponding AUC values, whose results are in the middle panel of Figure \ref{fig:inn_analysis}. 
%The AUC values of $T=100$ are slightly better than those of $T=200$ though the benefit is not significant.

\subsection{Loss function}
\label{sec:loss_function}

\begin{figure}[t]
\renewcommand\thefigure{E.1} 
\vskip 0.1in
\begin{center}
\centerline{
\includegraphics[width=0.3\textwidth]{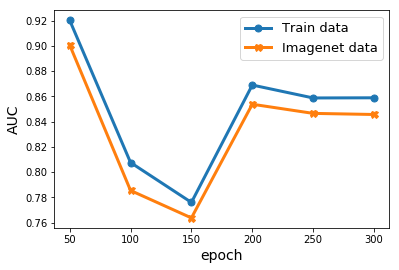}
\includegraphics[width=0.3\textwidth]{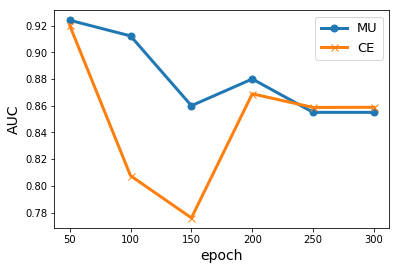}
}
\caption{AUC values for clean/noisy sample classification of the INN method obtained by varying two settings:  
%We vary three factors: 
({\bf Left}) training data set for learning $h$ 
%({\bf Middle}) training epoch for learning $h$, 
and ({\bf Right}) loss function for $f$. We analyze the 40\% asymmetrically noisy CIFAR10.
}
\label{fig:inn_analysis}
\end{center}
\vskip -0.1in
\end{figure}

We investigate two loss functions to estimate the prediction model $f(\cdot;\hat{\theta})$: 1) the standard cross-entropy (CE) function and 2) the MixUp (MU) objective function \citep{zhang2017mixup}. %and 3) the sum of CE and MU (CE+MU). 
By each 50 training epoch, we calculate their INN scores on the training data and evaluate their AUC values. 
The right panel in Figure \ref{fig:inn_analysis} shows that the AUC values derived by the MU function are higher and relatively stable. 
%Hence we choose the MU objective function as the loss function ($l$) of the prediction model $f(\cdot;\theta)$ unless otherwise stated.

%\subsection{The number of neighborhoods}
%\label{sec:num_neighbors}
%By each 50 training epoch, we calculate INN scores by varying the number of neighborhoods $L$ from one to 200, and evaluate the corresponding AUC values as Section \ref{sec:diss_msr} and \ref{sec:loss_function} did. 
%As can be seen in the lower right panel in Figure \ref{fig:inn_analysis},  the clean/noise classification performance of the INN scores becomes better as $L$ increases. 
%Note that the time complexity to calculate the INN score is a linear order of $L$, i.e. $O(L)$. And the rate of increase of AUC with respect to $L$ declines rapidly when $L>10$, thus we set $L$ to 10 in the numerical study to save computing times. 

%\newpage

\subsection{Training time analysis}

We analyze the total training time of the INN over CIFAR10 and compare it to the small-loss competitor. 
We use a single NVIDIA TITAN XP GPU and the results are summarized in Table \ref{table:train_time_inn} and \ref{table:train_time_loss}. 
The elapsed time of the INN is two times longer than that of the small-loss method since investigating the nearest neighborhoods requires much more time than other steps. %, such as training prediction models with a standard loss function. 
Thus, to overcome the limitation of the INN in terms of elapsed time, we need to modify and lighten the procedure to calculate neighborhoods.

\begin{table}[ht]
\fontsize{8.2pt}{8.2pt}
\selectfont
\renewcommand\thetable{E.1} 
\centering
\caption[9pt]{Total elapsed time for running the INN. We train two deep architectures ($f$ and $h$) for 50 epochs by minimizing the MixUp loss function. 
}
\label{table:train_time_inn}
\vskip 0.11in
\begin{center}
\begin{tabular}{c|c|c|c}

\hline
Training two architectures  & Searching the nearest neighbors & Calculating the INN scores&Total \\ 
    
\hline
\hline
21.5sec/ep$\times$50ep        & 2520.4sec        & 890.9sec     & 4486.3sec\\
\hline
\end{tabular}
\end{center}
\vskip -0.11in
\end{table}

\begin{table}[H]
\fontsize{8.2pt}{8.2pt}
\selectfont
\renewcommand\thetable{E.2} 
\centering
\caption[9pt]{Total elapsed time for running the small-loss strategy. We train a deep architecture for 50 epochs by minimizing the sum of the cross-entropy and negative entropy.
}
\label{table:train_time_loss}
\vskip 0.11in
\begin{center}
\begin{tabular}{c|c|c}

\hline
Training an architecture   & Calculating the loss scores & Total \\ 
    
\hline
\hline
40.2sec/ep$\times$50ep        &  3.8sec      & 2013.8sec\\
\hline
\end{tabular}
\end{center}
\vskip -0.11in
\end{table}

%\begin{table}[t]
%\fontsize{8.2pt}{8.2pt}
%\selectfont
%\renewcommand\thetable{E.3} 
%\centering
%\caption[9pt]{Comparison of the elapsed time(second). The results marked with * are re-implemented by us. 
%}
%\label{table:train_time2}
%\vskip 0.11in
%\begin{center}
%\begin{tabular}{l|c|c|c|c|c|c}

%\hline
%    & Step1.    & Nearest neighbor. & Step2. & INN &   SL   &    SSL  \\ 
    
%\hline
%\hline
%INN+DivideMix   &  3225.0(21.5)         & 2520.4         & 1070.0(21.4)  & 890.9     & -  & 27270(90.9)  \\
%\hline
%DivideMix*  & -      &   -    & -    & -  &  1206.0(40.2)    & 27390.0(91.3)     \\
%\hline
%\end{tabular}
%\end{center}
%\vskip -0.11in
%\end{table}

\end{document}